%% file: main.tex
\documentclass[runningheads]{llncs}

\usepackage{eccv}

\usepackage{eccvabbrv}

\usepackage{graphicx}
\usepackage{booktabs}
\usepackage{multirow}
\usepackage[table]{xcolor}

\usepackage[accsupp]{axessibility}  
\input{preamble}
\usepackage{hyperref}

\usepackage{orcidlink}

\makeatletter
\def\@fnsymbol#1{\ensuremath{\ifcase#1\or\star\or \dagger\or
   {\star\star\star}\or \dagger\or \ddagger\or
   \mathchar "278\or \mathchar "27B\or \|\or **\or \dagger\dagger
   \or \ddagger\ddagger \else\@ctrerr\fi}}
\makeatother

\begin{document}

\title{Gaussian-Voxel Duet: A Dual-Scaffolding Hybrid Representation for Fast and Accurate Monocular Surface Reconstruction}

\titlerunning{Gaussian-Voxel Duet}

\author{
Zhenhua Du\inst{1,2}\thanks{Equal contribution.} \and
Zhen Tan\inst{3}$^{\star}$ \and
Haoyu Zhang\inst{3} \and\\
Dewen Hu\inst{3} \and
Shuaifeng Zhi\inst{3}\thanks{Corresponding author.} \and
Peidong Liu\inst{2}
}

\authorrunning{Z. Du et al.}

\institute{
Zhejiang University \and
Westlake University \and
National University of Defense Technology
}

\maketitle

\input{sec/0_abstract}


\input{sec/1_intro}

\input{sec/2_related}
\input{sec/3_method}

\input{sec/4_experiment}
\input{sec/5_conclusion}

%
%
\bibliographystyle{splncs04}
\bibliography{refs}

\input{sec/X_suppl}

\end{document}

%% file: sec/0_abstract.tex
\begin{abstract}
While 3D Gaussian Splatting has achieved remarkable success in photorealistic novel view synthesis, its pursuit of fast and high-fidelity 3D reconstruction has long been constrained by a trade-off between geometric accuracy and optimization efficiency. Methods specialized in image rendering converge quickly at the cost of imperfect geometry caused by superfluous primitives overfitting training views, while methods integrating neural signed-distance field (SDF) for better geometry incur prohibitive training costs. In this paper, we attempt to strike a better trade-off by tethering scaffold-anchored Gaussians to a jointly optimized sparse voxel scaffold. This hybrid Gaussian-Voxel representation explicitly confines anchored Gaussians to a narrow band around surfaces defined by voxelized SDFs, which effectively improves representation efficiency and condenses floating Gaussians without sacrificing geometry quality. An implicit surface tethering loss further pulls individual Gaussian primitives closer to SDF-induced surfaces in a mutually regularized manner for improved reconstruction accuracy. Extensive experiments on diverse real-world indoor scenes from ScanNet++, ScanNetv2, and DeepBlending datasets demonstrate that our method achieves state-of-the-art surface reconstruction quality as well as superior novel view synthesis against leading baselines, while maintaining fast training convergence and real-time rendering. Code will be available at \url{https://github.com/duzh11/VoxelGS}.
  \keywords{Hybrid Scene Representation \and Surface Reconstruction \and View Synthesis \and 3d Deep Learning}
\end{abstract}

%% file: sec/1_intro.tex
\section{Introduction}
\label{sec:intro}

\begin{figure}[!ht]
    \centering
    \includegraphics[width=1\linewidth]{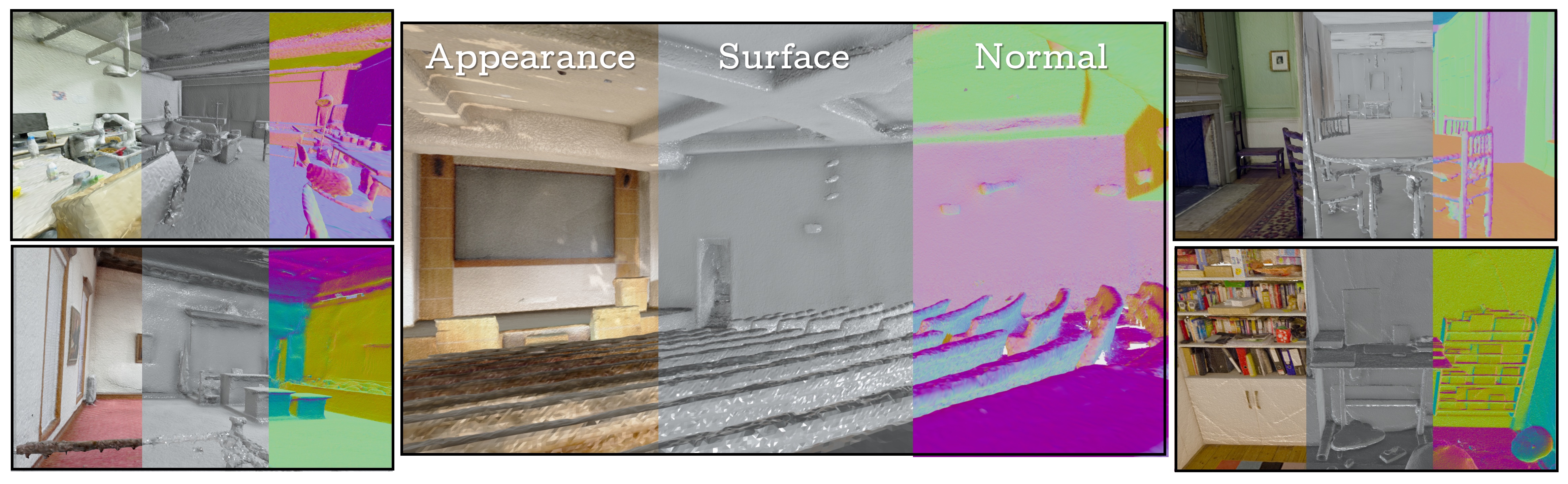}
    \caption{\textbf{Fast Monocular Surface Reconstruction with our Gaussian-Voxel Dual-Scaffolding Representation.} Our method, Gaussian-Voxel Duet, unleashes the structural advantages of anchored-Gaussians and sparse volumetric SDF field, producing high-fidelity meshes of challenging textureless and large-scale 3D indoor scenes. Our method achieves state-of-the-art performance while attaining compelling efficiency. }
    \label{fig:teaser}
\end{figure}

Multi-view 3D scene reconstruction remains a core challenge in computer vision and graphics. The emergence of Neural Radiance Fields (NeRF) \cite{NeRF} and its variants \cite{neus, volsdf, neuralangelo, monosdf} marked a paradigm shift, demonstrating remarkable fidelity in modeling complex appearance and fine geometry from posed RGB images. However, their reliance on dense, volumetric ray marching often renders them computationally prohibitive. The recent introduction of 3D Gaussian Splatting (3DGS) \cite{3DGS} represented a significant leap forward, by replacing costly volume rendering with efficient point-based splatting, achieving real-time rendering while preserving high visual quality. Despite its excellence in novel view synthesis, a critical limitation of 3DGS persists: the underlying geometry recovered by 3DGS and its derivatives is often inaccurate. This impedes reliable surface extraction and hinders downstream applications demanding precise geometric reasoning.

The root of this geometric inaccuracy lies in the shape-radiance ambiguity \cite{NeRF++}. Under monocular color supervision, 3DGS is prone to producing superfluous `floater' Gaussians that minimize photometric error for plausible rendering, yet lie far from the true surface. Existing mitigation strategies often involve establishing various structural scaffolds \cite{scaffold-gs} to constrain Gaussian primitives. One line of work \cite{3dgsr, GSDF, GS-Pull, discretized-sdf} learns a neural Signed Distance Function (SDF) to pull Gaussians toward an implicit surface. While it improves geometry quality, the integration of a global, neural SDF introduces substantial computational overhead, escalating per-scene optimization costs to several hours.
Another line of work chooses to anchor Gaussian primitives to explicit topological structures such as voxel grids or meshes \cite{svr, GeoSVR}. These methods preserve fast rendering speed and enhance surface coherence, but sacrifice the inherent flexibility of the Gaussian representation, causing geometric degradation in sparsely observed areas or flat and smooth structures. Moreover, their success often depends on robust geometric initialization or additional scene-level bounding-box priors, potentially limiting their broad applications. 
Therefore, one important question arises: \textit{How can we achieve both high geometric accuracy and computational efficiency in monocular surface reconstruction?}

In this paper, we propose a simple yet effective solution for fast and robust 3D reconstruction from multi-view monocular images. 
Our approach is built upon a novel dual-scaffold representation, which collaboratively learns a Gaussian-anchor scaffold and a voxel scaffold. 
The anchor scaffold, inspired by \cite{scaffold-gs}, structures Gaussian primitives in locally managed sets to reduce redundancy and improve rendering fidelity. 
The voxel scaffold, in contrast, captures the underlying geometry surface through a collection of differentiable sparse voxels, each encoding a local Signed Distance Function (SDF). In contrast to existing Gaussian-SDF hybrid representations \cite{GSDF,GS-Pull}, we avoid learning a costly global SDF from scratch by focusing on a sparse voxel grid covering scene surfaces and optimizing only residual SDF updates, which enables significantly accelerated convergence. 
By tethering scaffold-anchored Gaussians to a jointly optimized sparse voxel scaffold, we achieve a better trade-off among geometric accuracy, converging efficiency, and rendering quality.

The core of our framework (as shown in \cref{fig:teaser})  is a mutual tethering mechanism that tightly couples the two scaffolds. 
First, an explicit anchor tethering strategy uses the voxel scaffold's SDF to confine anchor densification and pruning to a narrow, surface-proximal confidence band, effectively eliminating geometric outliers (\cref{fig:pcd_comparison}). 
Second, we introduce an implicit surface tethering loss that pulls individual Gaussians toward the SDF-defined surface, simultaneously refining the Gaussian distribution and providing direct supervisory signals to update voxel SDF values. 
This bidirectional regularization allows Gaussians to adapt freely within a geometrically grounded region, preserving the high fidelity of novel view synthesis while ensuring surface coherence. 
We achieve high-fidelity geometry and appearance with fast training convergence and real-time rendering on complex and large indoor scenes.

In summary, our contributions are as follows:
\begin{itemize}
    \item 
    We propose a collaborative dual-scaffold reconstruction approach featuring a hybrid Gaussian–voxel representation that enables efficient high-quality surface reconstruction from posed multi-view images.

    \item We introduce a tethering mechanism, comprising an explicit anchor-tethering strategy and an implicit surface-tethering strategy, to ensure Gaussian primitives remain geometrically grounded without sacrificing flexibility and rendering quality.
    
    \item Extensive experiments on ScanNet++, ScanNetv2, and DeepBlending demonstrate that our method achieves state-of-the-art surface reconstruction and superior novel view synthesis against leading baselines, while maintaining fast training and real-time rendering.
\end{itemize}

%% file: sec/2_related.tex
\section{Related Work}
\label{sec:related}

\paragraph{Surface Reconstruction via Implicit Neural Field.}
Implicit neural fields represent a 3D scene as continuous functions learned by neural networks. 
The seminal work NeRF \cite{NeRF} parameterizes a radiance field with an MLP that maps spatial locations to volume density and view-dependent color. 
Subsequent NeRF variants aim to improve the quality of the extracted geometry. 
These approaches integrate occupancy grids \cite{unisurf} or SDFs \cite{neus,volsdf} into the differentiable volume rendering pipeline by re-parameterizing volume density, thus yielding a well-defined isosurface for mesh extraction. 
Subsequent efforts have attempted to enhance reconstruction quality through additional regularization strategies \cite{monosdf,neuris,manhattan-sdf}. 
Other methods \cite{neuralangelo,neurodin} integrate SDF-based representations with hash encodings and coarse-to-fine hierarchical optimization to accelerate convergence and increase surface detail recovery. 
Nevertheless, implicit neural field methods typically require hours of per-scene optimization, motivating the development of more efficient scene representations.

\paragraph{Surface Reconstruction via Gaussian Splatting.} 
3D Gaussian Splatting \cite{3DGS} represents a scene as a collection of explicit learnable Gaussian primitives, enabling real-time novel-view synthesis with markedly shorter optimization times. 
However, extracting high-fidelity surfaces from the discrete Gaussians remains a significant challenge, which has motivated several complementary lines of research: One line of work constrains the Gaussian ellipsoids to flatten surfels \cite{2DGS,gaussian-surfels,PGSR,quadratic-GS}, forcing the flattened primitives to align with local surface geometry. 
A second direction attempts to convert Gaussian representations into volumetric density \cite{sugar} or an opacity field \cite{GOF}. 
For example, some approaches couple Gaussians with a neural SDF, either jointly optimizing an SDF branch \cite{GSDF,3dgsr,gs-sdf} or using SDF gradients to pull Gaussian primitives toward the zero-level set \cite{GS-Pull}. Currently, external regularization strategies such as monocular geometric cues \cite{Dn-splatter,Gaussianroom,indoorgs,vcr-gaus}, multi-view consistency \cite{PGSR,gsurf}, or semantic priors \cite{atlasgs} are also being leveraged to stabilize reconstruction in challenging scenes.
Despite this rapid progress, most GS-based surface reconstruction methods either lack a well-defined surface constraint, making them susceptible to floating artifacts in empty space and low-texture regions. 

\paragraph{Surface Reconstruction via Grid-based Neural Field.}
Applying sparse voxel grids, octrees, and hash grids to neural rendering and 3D reconstruction has been extensively explored \cite{plenoctrees,plenoxels,instant-ngp,neuralangelo,ash,torch-ash,H2mapping}, where spatial sparsity and structured storage enable empty-space skipping, fast feature queries, and tighter geometric control. Introducing grid hierarchies into Gaussian splatting has also proved effective: Scaffold-GS \cite{scaffold-gs} and Octree-GS \cite{octree-gs} regulate the spatial layout of Gaussians with gridified anchors to improve rendering stability and scalability. Furthermore, SVRaster \cite{svr} directly treats voxels as the rendering primitives for real-time high-fidelity radiance fields, and GeoSVR \cite{GeoSVR} augments this pipeline with geometric constraints to obtain accurate, mesh-extractable surfaces.

%% file: sec/3_method.tex
\begin{figure}[!ht]
    \centering
    \includegraphics[width=1\linewidth]{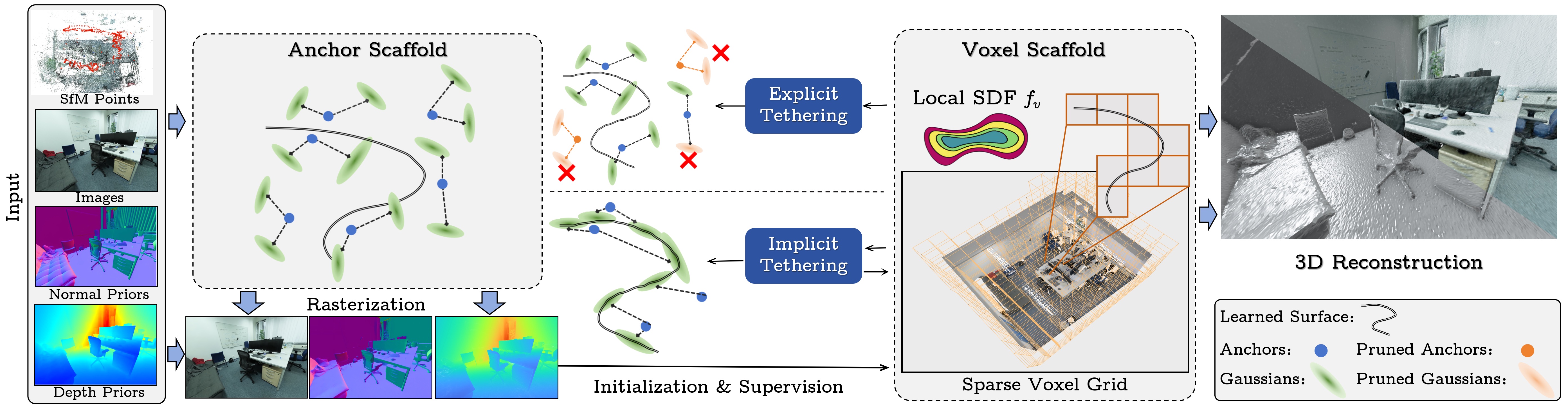}
    \caption{\textbf{Method Overview. }
Starting from multi-view images, SfM points, and monocular priors, we (1) build \textbf{a dual-scaffold hybrid representation}, where the anchor scaffold produces 2D Gaussian surfels for appearance and the voxel scaffold encodes a sparse local SDF for surface geometry, (2) perform \textbf{explicit tethering} to prune off-surface anchors and Gaussians based on the learned SDF, and (3) apply \textbf{implicit tethering} to further pull remaining Gaussians toward the learned surface for accurate 3D reconstruction.}
    \label{fig:overview}
\end{figure}

\section{Method}
\label{sec:method}
We present our dual-scaffold hybrid representation with jointly optimized Gaussian primitives and sparse voxel grids for scene reconstruction, as shown in \cref{fig:overview}.

Given multi-view posed images and imperfect monocular priors, an anchor scaffold tethering a set of 2D Gaussians is constructed via minimizing rendering loss. Concurrently, another sparse differentiable voxel scaffold is allocated along Gaussian-estimated surfaces and optimized to acquire accurate local signed distance fields (Sec. \ref{sec:method-1}).  We achieve high-quality geometry with efficient convergence by mutual regularization between these two scaffolds. Specifically, an explicit anchor-tethering strategy prunes distant anchored Gaussians while preserving those in proximity to the surface (Sec. \ref{sec:method-2}), and an implicit surface tethering strategy further pulls the remaining Gaussians and surfaces closer via joint optimization to ensure precise geometric and photometric quality (Sec. \ref{sec:method-3}).

\subsection{Preliminary}
\label{sec:preliminary}

\textbf{Gaussian Splatting}. 3D Gaussian Splatting (3DGS) \cite{3DGS} represents scenes using explicit 3D Gaussian ellipsoids $\{ \mathcal{G}_i \}$, each defined by its mean position $\boldsymbol{\mu}$ and covariance matrix $\boldsymbol{\Sigma}$: $\mathcal{G}(\boldsymbol{x}) = \exp\left( -\frac{1}{2}(\boldsymbol{x}-\boldsymbol{\mu})^{\top} \boldsymbol{\Sigma}^{-1} (\boldsymbol{x}-\boldsymbol{\mu}) \right)$,
where $\boldsymbol{x}$ denotes a 3D world coordinate. 3D Gaussians are projected to screen space as 2D Gaussians $\{ \mathcal{G}_i^{2D} \}$, with colors composited via: $\boldsymbol{c} = \sum_{i=1}^N \boldsymbol{c}_i \alpha_i \mathcal{G}_i^{2D} \prod_{j=1}^{i-1}\!\left(1 - \alpha_j \mathcal{G}_j^{2D}\right)$, where $\alpha_i$ is opacity and $\boldsymbol{c}_i$ is view-dependent color. Scaffold-GS \cite{scaffold-gs} further improves its rendering accuracy and robustness with anchor-spawned 3D Gaussians.

Although 3DGS and its variants excel at photorealistic image rendering, they lack a well-defined surface and hinder geometry extraction. To better align Gaussian primitives towards scene surfaces, 2D Gaussian Splatting (2DGS) opts for 2D Gaussian surfels defined on local tangent planes. Each surfel $\mathcal{G}_i$ is parameterized by center $\boldsymbol{p}_i$, tangent axes $\boldsymbol{t}_u, \boldsymbol{t}_v$, and scales $(s_u, s_v)$ in a local plane:
$P(u,v) = \boldsymbol{p}_i + s_u \boldsymbol{t}_u u + s_v \boldsymbol{t}_v v = \boldsymbol{H}(u, v, 1, 1)^T,$
where $\boldsymbol{H}$ is a $4\times4$ homogeneous transformation matrix representing 2D Gaussian geometry. 
In the local tangent coordinates $(u,v)$, the Gaussian surfel is defined as: $\mathcal{G}(u,v) = \exp\!\left(-\frac{u^2+v^2}{2}\right).$ Each 2D surfel is subsequently projected to screen space and follows a compositing procedure similar to 3DGS. In this paper, we also integrate the anchor scaffold into 2DGS to improve structural coherence and rendering quality.

\noindent\textbf{Neural Signed Distance Field.} Neural SDF represents scene geometry as a neural function $f$ mapping any 3D point $\boldsymbol{x}$ to its signed distance $s$ towards the nearest surfaces: $s = f_{d}(\boldsymbol{x})$. Thus, the scene surface is defined as the zero-level set of its SDF:
$\mathcal{S} = \{\, \boldsymbol{x} \in \mathbb{R}^3 \mid f(\boldsymbol{x}) = 0 \,\}$.

Existing neural SDF-based reconstruction approaches \cite{volsdf,neus, GSDF} typically apply a differentiable transformation that converts SDF value $s$ into opacity value or volume density, and enables end-to-end gradient propagation from photometric loss supervision. However, such a learning process that regresses an underlying dense SDF field from scratch generally exhibits slow convergence, requiring several hours for per-scene optimization.

\subsection{Dual-Scaffold Hybrid Representation}
\label{sec:method-1}




\noindent\textbf{Anchor Scaffold Representation.} 
Optimization of unstructured 3D Gaussian primitives often leads to geometric artifacts in large scenes, as there are excessive Gaussians drifting freely to minimize photometric error. However, simply fixing Gaussian positions would sacrifice the expressive view-dependent appearance they can model.

Inspired by Scaffold-GS \cite{scaffold-gs}, we introduce anchor scaffold to 2DGS for view-consistent planar Gaussian primitives as well as improved photometric rendering quality \cite{2DGS}, especially in complex large 3D scenes. Specifically, with anchors initialized from SfM points, each anchor then dynamically manages $k$ 2D Gaussian primitives $\{\mathcal{G}_0, \mathcal{G}_1, \dots, \mathcal{G}_{k-1}\}$, predicting their attributes conditioned on viewing parameters. Following previous monocular reconstruction approaches \cite{GeoSVR,monosdf,ash}, we also resort to monocular geometric cues of depths and normals \cite{metric3d,Stablenormal} for efficient warm-up of the training phase. Despite imperfect and view-inconsistent monocular priors, we experimentally found that our anchored 2DGS can leverage them to render view-consistent geometry after a few iterations ($\leq$ 1 min).

\begin{figure}[!t]
    \centering
    \includegraphics[width=1\linewidth]{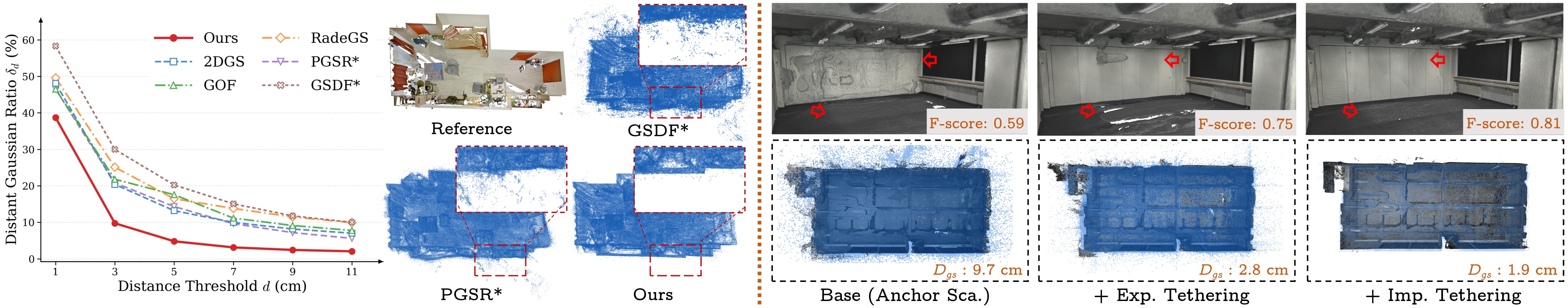}
    \caption{\textbf{Analysis of Gaussian Point Distributions}.
     \textbf{(Left)} We report the distant Gaussian ratio $\delta_d$ under varying thresholds $d$, alongside qualitative visualizations. Both quantitative metrics and qualitative comparisons demonstrate that our method effectively suppresses noisy floating Gaussians, yielding cleaner and more geometry-faithful distributions. \textbf{(Right)} On an extremely textureless scene (9460c8889d from ScanNet++), our explicit and implicit tethering strategies effectively align Gaussian primitives (blue points) toward the underlying GT surface, producing complete and accurate reconstruction. The metric $D_{gs}$ computes the mean distance between all Gaussian points and the GT mesh.}
    \label{fig:pcd_comparison}
\end{figure}

Our design choice of anchor scaffold significantly reduces the redundancy of Gaussians and regularizes their spatial distribution. We also notice that during optimization, the anchor scaffold itself lacks further geometric constraints but only obeys heuristic management \cite{scaffold-gs}, preventing strict alignment between anchored Gaussians and underlying geometry (\cref{fig:pcd_comparison}), which will be further addressed by our mutual optimization strategy in \cref{sec:method-2} and \cref{sec:method-3}.

\noindent\textbf{Voxel Scaffold Representation.} To promote explicit surface awareness into anchor scaffold representation, we introduce a voxel scaffold that represents the scene surface through a collection of sparse voxels allocated near object surfaces, each encoding a local signed distance $f_v(\mathbf{x})$ and RGB color $\boldsymbol{c}_v$.  Unlike methods that learn a global dense SDF field covering the entire scene, our approach focuses on learning local SDFs $f_v(\boldsymbol{x})$ near surfaces and only optimizes their residual updates. This strategy of local geometric modeling and residual learning enables detailed surface reconstruction and fast convergence, which in turn regularizes concurrent Gaussian distributions within the anchor scaffold.

However, in monocular reconstruction, initializing sparse voxels near the true surface presents a chicken-and-egg dilemma: accurate voxel placement requires surface knowledge, which is initially unavailable. We overcome this by creating sparse voxels using TSDF fusion of depth maps rendered by warmed-up Gaussians, and placing sparse voxels in estimated surface-proximal regions.



The above process establishes a core synergistic cycle: the voxel scaffold provides a geometrically accurate surface to which the anchor scaffold's Gaussians are guided, while the anchor scaffold, in turn, supplies the necessary supervisory geometry signals to update the voxel scaffold.

\subsection{Explicit Anchor Tethering}
\label{sec:method-2}
To ensure comprehensive coverage of scene surfaces, adaptive densification and pruning of anchors are essential. Prior work \cite{3DGS,scaffold-gs} follows a heuristic strategy based on positional gradients. Anchors exhibiting substantial accumulated gradients $\nabla_i$ are subdivided ($\nabla_i > \tau_g$), while those associated with persistently low-opacity Gaussians ($O_i$) are culled ($O_i < \tau_p$),
where $\tau_g$ and $\tau_p$ represent predetermined threshold hyperparameters. Despite effectively distributing anchors across the scene, it also produces unreliable anchors due to its reliance on dominant photometric cues and the objective for novel view synthesis, which cause anchors to remain geometrically ungrounded and lead to geometry artifacts in challenging areas like textureless or sparse-view regions.

We take advantage of our robust voxel scaffold to tether Gaussian anchors. Specifically, for an anchor $\mathcal{A}_i$ at $\boldsymbol{x}_i$, we define a confidence band $\mathcal{B} = \{ \boldsymbol{x} \in \mathbb{R}^3 \mid |f_v(\boldsymbol{x})| < \tau_d \}$ from the SDF $f_v$, which delineates the valid region for the anchor's placement and presence. 
We augment the tethering mechanism with additional geometric priors:
\begin{equation}
\begin{aligned}
&\text{Grow if:}\quad (\nabla_i > \tau_g) \land (|f_v(\boldsymbol{x}_i)| < \tau_d), \\
&\text{Prune if:}\quad (O_i < \tau_p) \lor (|f_v(\boldsymbol{x}_i)| > \tau_d),
\end{aligned}
\label{eq:exp-tether}
\end{equation}
where $\tau_d$ is a predefined threshold and controls the band's width. This ensures that densification is strictly applied to photometrically significant anchors near the surface, while the pruning operation removes optically inactive Gaussians and those lying far from the inferred surface. Additionally, we explicitly remove any anchor-spawned neural Gaussians residing beyond the valid boundary of the voxel scaffold, as they are deemed to be too distant from the surface.

By explicitly tethering anchors to the voxel scaffold, our approach enforces geometric coherence while preserving the flexibility needed for high-quality appearance modeling. In contrast to previous methods that introduce an SDF-aware pruning operator \cite{GSDF}, or rigidly attach Gaussians to voxel grids or mesh faces \cite{GeoSVR,meshgs}, our strategy balances simplicity, stability and adaptability, leading to more accurate geometry and improved novel view synthesis.

\subsection{Implicit Surface Tethering}
\label{sec:method-3}
The explicit anchor tethering strategy in Sec.~\ref{sec:method-2} has ensured anchors' proximity to the surface, but it does not directly constrain anchor-spawned Gaussian primitives. To close this gap, we introduce an implicit surface tethering loss that further pulls each remaining Gaussian toward its nearest surface by minimizing the tethering length, i.e., SDF value.

Formally, for a Gaussian primitive located at $\boldsymbol{p}_k$, we define the loss as:
\begin{equation}
    \mathcal{L}_{\text{t}} = \sum_{k=1}^N \Vert f_v(\boldsymbol{p}_k) \Vert_2,
    \label{eq:implicit-loss}
\end{equation}
where $f_v(\cdot)$ is the SDF evaluated from voxel scaffold. We adopt the $L_2$ loss which strongly penalizes Gaussians with large absolute SDF values (i.e., those relatively far from the surface), while granting greater freedom of movement to those already within the vicinity of the zero-level set.

This loss serves a dual purpose. First, it acts as a continuous attractor, pulling Gaussians toward the inferred geometry and refining their spatial distribution. Second, it also provides direct supervisory signals to the SDF field $f_v(\cdot)$ itself. The gradients flowing from $\mathcal{L}_{\text{t}}$ help update the queried SDF value, making voxel scaffold a more accurate geometric representation. As evidenced by our experiments, this simple yet effective loss facilitates a synergistic co-optimization between the anchor scaffold and the voxel scaffold, leading to improved geometric accuracy without compromising rendering fidelity.

\subsection{Implementation Details}
\label{sec:method-4}

\noindent\textbf{Anchor scaffold optimization.}
Following~\cite{2DGS, scaffold-gs}, we optimize the anchor scaffold with standard photometric loss $\mathcal{L}_c$, depth distortion loss $\mathcal{L}_d$, and normal consistency loss $\mathcal{L}_n$. Monocular depth and normal cues ~\cite{metric3d,Stablenormal} are incorporated by formulating corresponding losses $\mathcal{L}_{md}$ and $\mathcal{L}_{mn}$ as a Pearson depth loss and an $L_1$ loss, respectively. The overall anchor–Gaussian objective is 
\begin{equation}
    \mathcal{L}_{gs}
    = \mathcal{L}_c
    + \lambda_d \mathcal{L}_d
    + \lambda_n \mathcal{L}_n
    + \lambda_{md} \mathcal{L}_{md}
    + \lambda_{mn} \mathcal{L}_{mn},
    \label{eq:loss_gs}
\end{equation} 
where the weights $\lambda_{md}$ and $\lambda_{mn}$ follow an annealing schedule with decreasing dependence on monocular priors.

\noindent\textbf{Voxel scaffold optimization.}
We implement our voxel scaffold using the differentiable sparse voxel engine of Torch-Ash~\cite{torch-ash}. After warming up the anchor scaffold for $N_{w}$ iterations, we unproject the Gaussian-rendered depth to initialize a sparse voxel grid. The voxel SDF, color, and normal are then optimized with a voxel color loss $\mathcal{L}_{vc}$, voxel depth loss $\mathcal{L}_{vd}$, voxel normal loss $\mathcal{L}_{vn}$, and an Eikonal regularizer $\mathcal{L}_{eik}$:
\begin{equation}
    \mathcal{L}_{v}
    = \mathcal{L}_{vc}
    + \lambda_{vd} \mathcal{L}_{vd}
    + \lambda_{vn} \mathcal{L}_{vn}
    + \lambda_{eik} \mathcal{L}_{eik}.
    \label{eq:loss_voxel}
\end{equation} 
In practice, we set the warm-up iteration $N_{w}\in\{1500, 3000\}$ depending on the scene scale, and reset the voxel scaffold every $1500$ iterations to mitigate the influence of outdated initialization. 

\noindent\textbf{Overall loss function.}
The proposed tethering loss $\mathcal{L}_t$ in Sec.~\ref{sec:method-3} links Gaussians to the voxel scaffold and confines them to a narrow surface band. The overall training objective is:
\begin{equation}
    \mathcal{L}
    = \mathcal{L}_{gs}
    + \mathcal{L}_{v}
    + \lambda_t \mathcal{L}_{t},
\label{eq:loss_overall}
\end{equation}
where $\lambda_t$ controls the strength of tethering.  Most of the hyper-parameters, including loss weights and optimization settings, are set following 2DGS~\cite{2DGS}, Scaffold-GS~\cite{scaffold-gs}, and Torch-Ash~\cite{torch-ash}, and we found our performance insensitive to the particular choice of hyper-parameters. More details can be seen in Appendix.

%% file: sec/4_experiment.tex
\section{Experiment}
\label{sec:experiment}

\begin{figure*}[!t]
    \centering
    \includegraphics[width=1\linewidth]{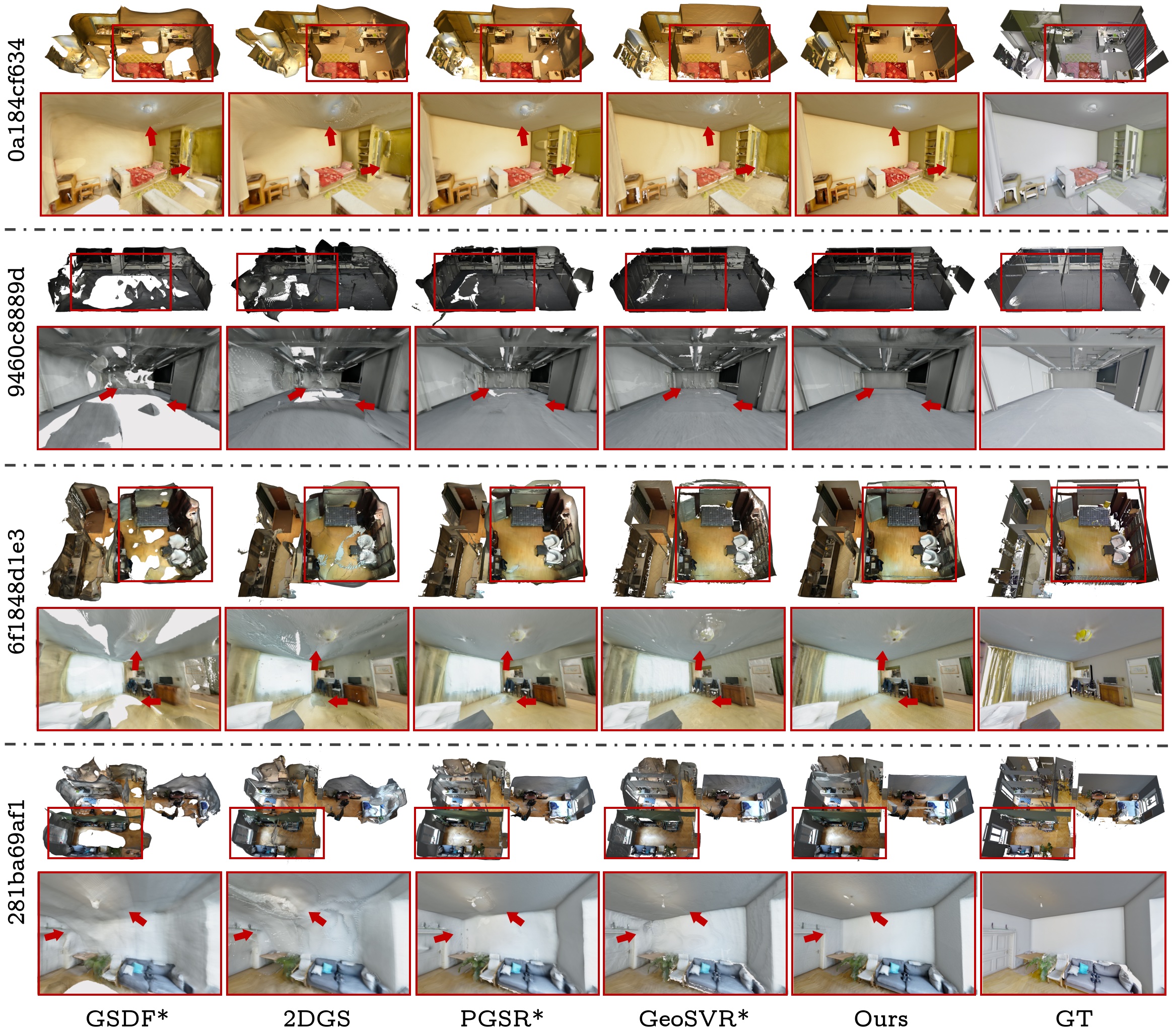}
    \caption{\textbf{Qualitative Results of Surface Reconstruction on ScanNet++}.
    We show the global mesh (top) and detailed view of the region indicated by the red box (bottom) of four scenes. Compared to 2DGS and monocular-enhanced baselines (GSDF*, PGSR*, GeoSVR*), our approach achieves the best reconstruction quality. Specifically, our method yields cleaner geometry with fewer floating artifacts and smoother surfaces, especially in texture-less regions and large-scale multi-room layouts.}
    \label{fig:surface_recon_scannetpp}
\end{figure*}

\subsection{Experimental Setup}
\noindent \textbf{Datasets.} We evaluate our method on three real-world indoor datasets: ScanNet++ \cite{scannet++}, ScanNetv2 \cite{scannet}, and DeepBlending \cite{DeepBlending}. For ScanNet++, we randomly select 12 scenes covering various spatial scales and layout complexity. For ScanNetv2 and DeepBlending, we follow the same scene selections as MonoSDF~\cite{monosdf} and 3DGS~\cite{3DGS}, respectively.

\noindent\textbf{Metrics.} For surface reconstruction, we strictly follow the evaluation protocol of~\cite{atas} and report five geometric metrics: accuracy (Acc), completeness (Comp), precision (Prec), recall (Recall), and F-score. For novel view synthesis, we adopt standard image quality metrics: PSNR, SSIM, and LPIPS.

\noindent\textbf{Baselines.}
We compare our method against the following representative reconstruction baselines:
(1) implicit methods: MonoSDF~\cite{monosdf}, and Ash~\cite{ash};
(2) explicit methods: 2DGS~\cite{2DGS}, RaDeGS~\cite{Rade-GS}, GOF~\cite{GOF}, PGSR~\cite{PGSR}, and GeoSVR~\cite{GeoSVR};
(3) hybrid methods: GSDF~\cite{GSDF} and GS-Pull~\cite{GS-Pull}.
To fairly compare against strong reconstruction baselines with monocular priors, we equip PGSR and GSDF with the same depth and normal supervision (denoted as PGSR* and GSDF*, respectively), and add monocular normal cues to GeoSVR (denoted as GeoSVR*), leading to a further performance boost.

\subsection{Surface Reconstruction}
\begin{figure*}[!t]
    \centering
    \includegraphics[width=1\linewidth]{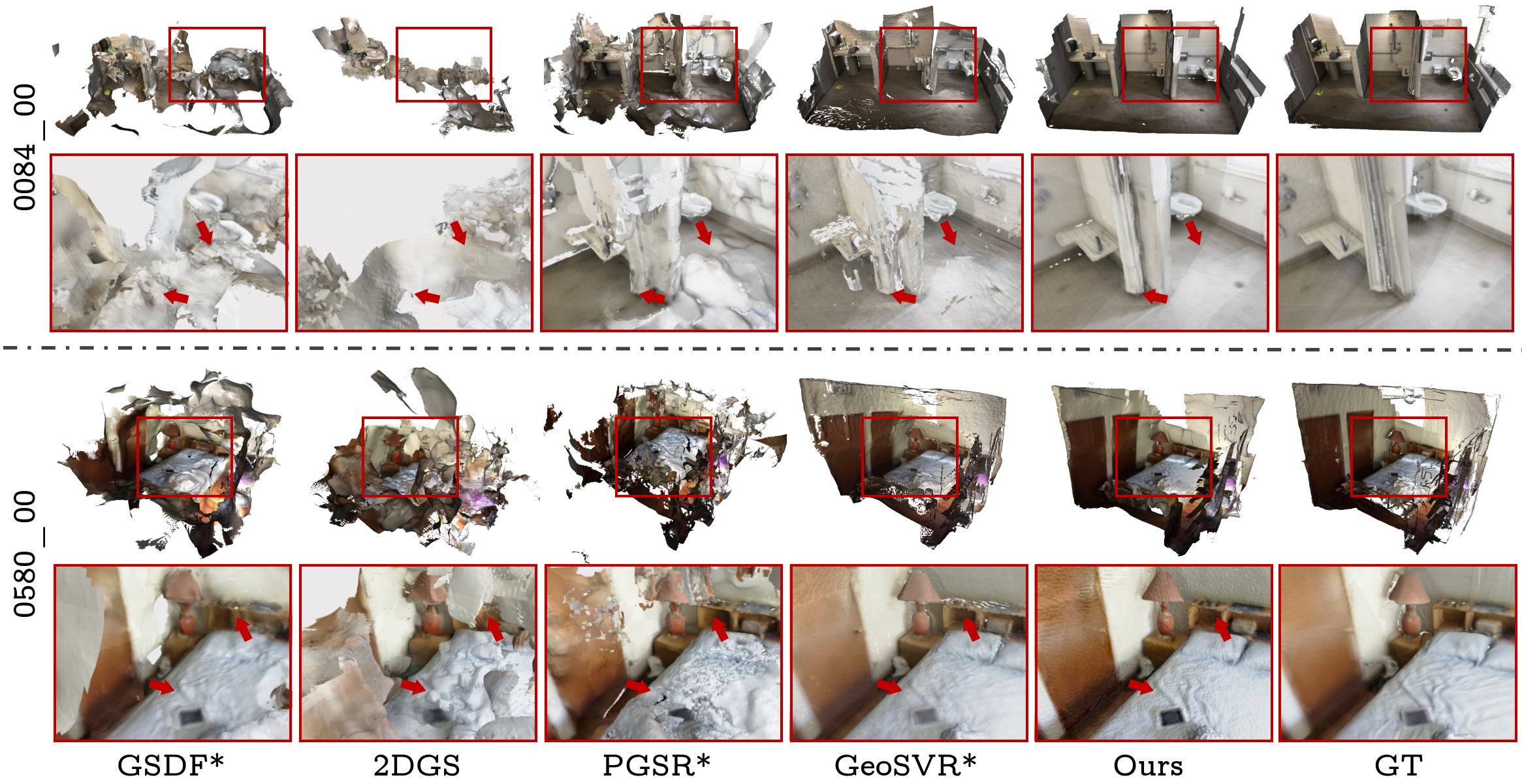}
    \caption{\textbf{Qualitative Results of Surface Reconstruction on ScanNetv2}.
    We visualize the global mesh (top) and zoomed-in views (bottom) of the highlighted regions. Compared to 2DGS and prior-aided baselines (GSDF*, PGSR*, GeoSVR*), our method consistently yields cleaner geometry with sharper details and fewer artifacts.}
    \label{fig:surface_recon_scannet}
\end{figure*}

\input{tab/surface}
We conducted extensive quantitative comparisons on the popular ScanNet++ and ScanNetv2 datasets, as shown in \cref{tab:surface_recon}. On the ScanNet++ dataset, our method outperforms all baselines, achieving the highest F-score, recall, and precision. This clearly surpasses SOTA surface reconstruction methods like GeoSVR and PGSR with fair monocular cues. Qualitative results in \cref{fig:surface_recon_scannetpp} demonstrate our superior surface reconstruction quality on both challenging textureless room-scale scenes and multi-room large-scale scenes. Our method produces precise and smooth meshes without the structural artifacts observed in the baselines. \cref{fig:pcd_comparison} also shows that our method manages to prune distant Gaussians and leaves significantly more Gaussian primitives lying close to the surface.

On the ScanNetv2 dataset, our method also consistently performs the best in F-score, outperforming SOTA NeRF-based implicit and voxel-based explicit reconstruction methods MonoSDF and GeoSVR*, both of which have access to depth and normal priors. 
Notably, explicit baselines struggle with ScanNetv2's noisy COLMAP initialization and motion blur unlike on the high-fidelity ScanNet++ (See Appendix C.1 for more details). This issue also undermines hybrid approaches that bootstrap SDF learning from optimized Gaussians like GSDF and GS-Pull. By explicitly deleting distant Gaussians and anchoring them to the learned SDF surface, our method effectively prevents drift under noisy initialization, enabling robust reconstruction. \cref{fig:surface_recon_scannet} further validates the robustness of our method when processing images captured with motion blur and illumination variation. Our extracted 3D meshes are visually better structured with smooth surfaces and fewer artifacts.

\input{tab/nvs}

\subsection{Appearance Reconstruction}
\begin{figure*}[!t]
    \centering
    \includegraphics[width=1\linewidth]{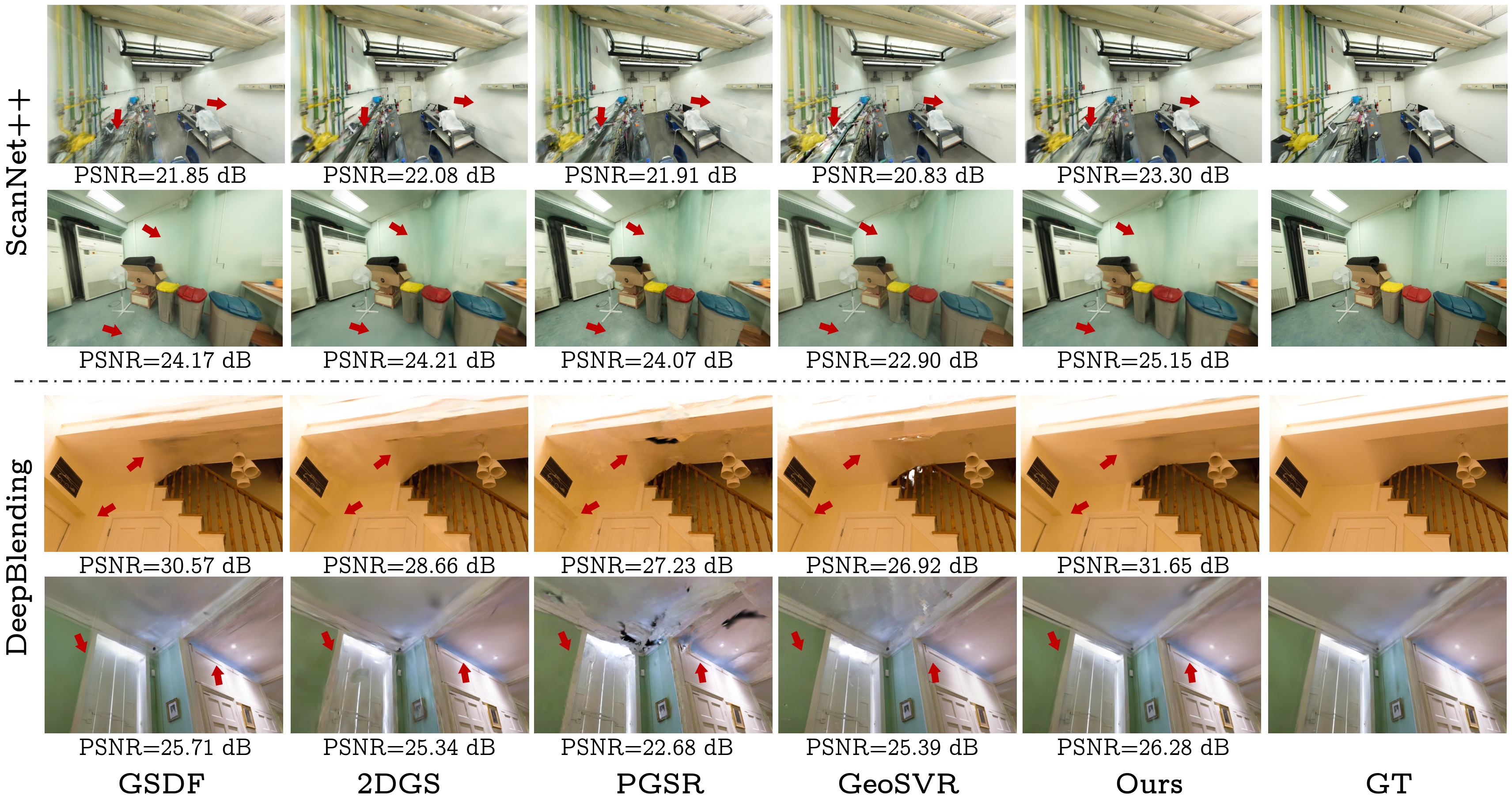}
    \caption{\textbf{Qualitative Results of NVS}. We visualize the NVS results on ScanNet++ and DeepBlending scenes, respectively. While baselines suffer from severe ghosting and artifacts, our method consistently achieves superior rendering quality and robustness.}
    \label{fig:nvs}
\end{figure*}

\begin{figure*}[!t]
    \centering
    \includegraphics[width=1\linewidth]{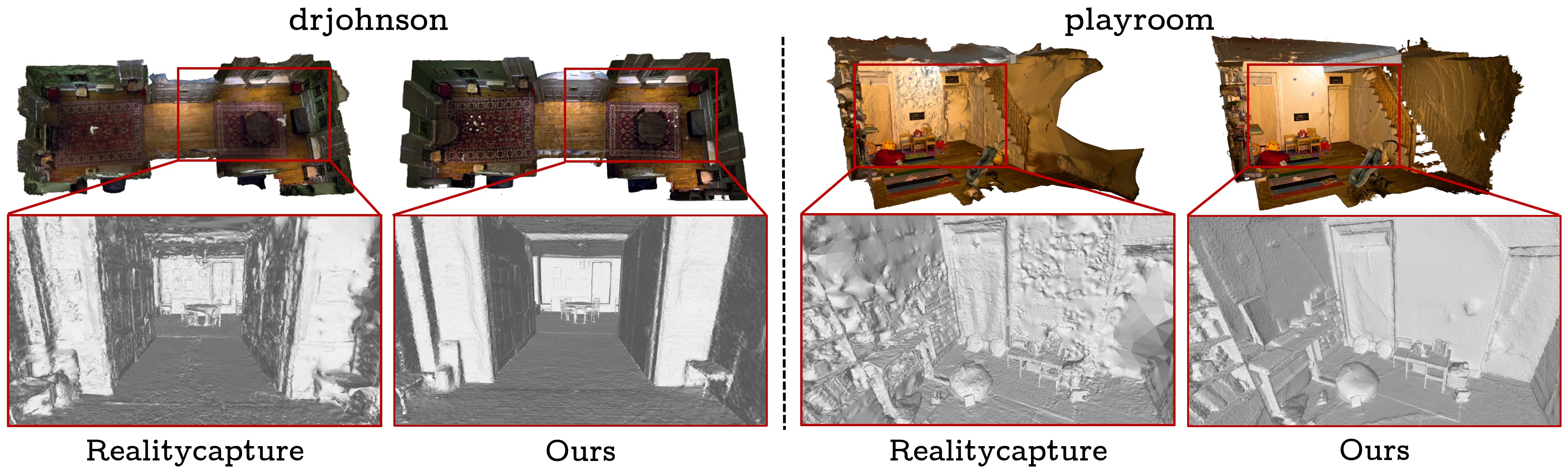}
    \caption{\textbf{Qualitative Results of Surface Reconstruction on DeepBlending}. We compare the mesh produced by our method with the GT mesh provided in the official dataset (generated by RealityCapture). Our method recovers faithful geometry and high-fidelity textures comparable to the photogrammetry baseline.}
    \label{fig:surface_recon_db}
\end{figure*}

Though not exclusively designed for photorealistic novel view synthesis, in addition to accurate surface reconstruction, our method also achieves high-quality image rendering. 
We tested the NVS performance on two different setups: ScanNet++ focuses on view extrapolation since its testing views deviate from training-view trajectories, while DeepBlending focuses on the view interpolation setup. 
As shown in \cref{tab:nvs}, our method achieves the highest PSNR on the ScanNet++ dataset and competitive rendering quality on the DeepBlending dataset. Qualitative NVS results are shown in \cref{fig:nvs}, and textured meshes for DeepBlending scenes are presented in \cref{fig:surface_recon_db}. Since our reconstructions exhibit better geometric fidelity than the RealityCapture reference meshes, we do not report quantitative reconstruction metrics on DeepBlending.

\subsection{Efficiency Analysis}


With the dual-scaffold design, our method learns a local SDF field and only restricts Gaussians close to the surfaces. 
As a result, we substantially improve convergence efficiency compared to SOTA Gaussian-SDF method GSDF ($9\times$ faster) \cite{GSDF} as well as the explicit method GeoSVR \cite{GeoSVR}, as shown in \cref{tab:efficiency}. Note that all methods are evaluated on a single RTX 4090 GPU, except that GSDF is evaluated on an H20 GPU due to Out-of-Memory issue.


\subsection{Ablation Study}
\input{tab/efficiency}
\label{sec:abla}

\input{tab/ablation}

In this section, we systematically verify the effectiveness of our key design choices in high-quality reconstruction, using the 12 randomly selected scenes from ScanNet++. 
In addition to metrics of precision, recall and F-score which concern geometry quality within a pre-defined threshold (typically set to 5cm), we conduct a further investigation by measuring the average absolute distance between Gaussian positions and ground-truth surface (denoted as $D_{gs}$(cm)) as well as the ratio of `outlier` Gaussian points deviating from ground-truth surface beyond 5cm  (denoted as $\delta_{5}$(\%)), aiming to provide a more comprehensive view of underlying reconstruction quality. 
\cref{tab:ablation} reports the contribution of voxel scaffold (Voxel Sca.), explicit anchor tethering (Exp.), and implicit surface tethering (Imp.) when successively applied to our 2DGS anchor scaffold (Anchor Sca.). 
We observe a clear and considerable performance boost from each of our key modules. Our full model reaches an average Gaussian-surface distance of 2.6cm and has only 4.81\% of Gaussians more than 5cm away from the ground truth surface. These results also validate our high F-score in \cref{tab:surface_recon}. Qualitative comparisons are shown in \cref{fig:pcd_comparison}.

We have further explored the ratio of `outlier' Gaussian points given different thresholding distances $d$ ranging from 1cm to 11cm, as shown in \cref{fig:pcd_comparison}. Compared to all popular Gaussian-based approaches, our method consistently has far fewer distant Gaussians under various thresholding distances, with more than 90\% of Gaussians located within 3cm of the underlying geometric surface.

%% file: tab/surface.tex
\begin{table*}[!t]
\centering
\renewcommand{\arraystretch}{1.2}
\centering
\caption{\textbf{Quantitative Results of Surface Reconstruction on ScanNet++ and ScanNetv2}. \colorbox{red!25}{First}, \colorbox{orange!25}{second}, and \colorbox{yellow!25}{third} best results are highlighted. * means methods enhanced with monocular cues. The symbol \dag~indicates the results on ScanNetv2 datasets cited directly from the original papers.}
\label{tab:surface_recon}
\resizebox{\linewidth}{!}{
\begin{tabular}{c l c c c c c c c c c c}
\cmidrule[\heavyrulewidth]{1-12}
\multirow{2}{*}{} & \multirow{2}{*}{\textbf{Method}} & \multicolumn{5}{c}{\textbf{ScanNet++}} & \multicolumn{5}{c}{\textbf{ScanNetv2}} \\
\cmidrule(r){3-7} \cmidrule(l){8-12}
 & & {Acc}$\downarrow$ & {Comp}$\downarrow$ & {Prec}$\uparrow$ & {Recall}$\uparrow$ & {F-score}$\uparrow$ & {Acc}$\downarrow$ & {Comp}$\downarrow$ & {Prec}$\uparrow$ & {Recall}$\uparrow$ & {F-Score}$\uparrow$\\ 
\cmidrule[\heavyrulewidth]{1-12}

\multirow{2}{*}{\rotatebox[origin=c]{90}{{Impl.}}} 
 & MonoSDF\dag \ \cite{monosdf}       & 0.077     & 0.082     & 0.566     & 0.494     & 0.524     & \colorbox{red!25}{0.035}      & \colorbox{orange!25}{0.048}      & \colorbox{red!25}{0.799}      & \colorbox{orange!25}{0.681}      & \colorbox{orange!25}{0.733} \\
 & Ash\dag \ \cite{torch-ash}         & 0.235 & 0.065 & 0.336 & 0.518 & 0.400 & \colorbox{orange!25}{0.042} & \colorbox{yellow!25}{0.056} & \colorbox{yellow!25}{0.751} & \colorbox{yellow!25}{0.678} & \colorbox{yellow!25}{0.710} \\
\midrule
\multirow{7}{*}{\rotatebox[origin=c]{90}{{Expl.}}} 
 & 2DGS \cite{2DGS}          & 0.122   & 0.084   & 0.583   & 0.607   & 0.593   & 0.262 & 0.339 & 0.183 & 0.171 & 0.173 \\
 & GOF \cite{GOF}           & 0.188   & 0.064   & 0.488   & 0.618   & 0.541   & 0.194 & 0.267 & 0.253 & 0.255 & 0.252 \\
 & RadeGS \cite{Rade-GS}     & 0.165   & 0.057   & 0.471   & 0.659   & 0.543   & 0.179 & 0.330 & 0.319 & 0.247 & 0.276 \\
 & PGSR \cite{PGSR}          & 0.157   & 0.065   & 0.640   & 0.687   & 0.660   & 0.272 & 0.167 & 0.319 & 0.342 & 0.330 \\
 & PGSR* & 0.081   & 0.047   & 0.718   & 0.732   & 0.723   & 0.127 & 0.141 & 0.396 & 0.372 & 0.383 \\
 & GeoSVR \cite{GeoSVR}      & \colorbox{orange!25}{0.047} & \colorbox{yellow!25}{0.045} & \colorbox{yellow!25}{0.783} & \colorbox{yellow!25}{0.774} & \colorbox{yellow!25}{0.778} & 0.066 & 0.068 & 0.551 & 0.527 & 0.538 \\
 & GeoSVR* & \colorbox{red!25}{0.046} & \colorbox{orange!25}{0.044} & \colorbox{orange!25}{0.788} & \colorbox{orange!25}{0.778} & \colorbox{orange!25}{0.782} & 0.061 & 0.064 & 0.575 & 0.551 & 0.562 \\
\midrule
\multirow{3}{*}{\rotatebox[origin=c]{90}{{Hybr.}}} 
 & GSDF \cite{GSDF}          & 0.182   & 0.405   & 0.508   & 0.391   & 0.436   & 0.160 & 0.207 & 0.264 & 0.182 & 0.215 \\
 & GSDF*                     & 0.071 & 0.113 & 0.612 & 0.516 & 0.559 & 0.158 & 0.161 & 0.318 & 0.257 & 0.283  \\
 & GS-Pull \cite{GS-Pull}    & 0.240   & 0.155   & 0.377   & 0.348   & 0.356   & 0.200 & 0.194 & 0.251 & 0.226 & 0.236 \\
 & \textbf{Ours}            & \colorbox{yellow!25}{0.076} & \colorbox{red!25}{0.027} & \colorbox{red!25}{0.805} & \colorbox{red!25}{0.890} & \colorbox{red!25}{0.842} & \colorbox{yellow!25}{0.055} & \colorbox{red!25}{0.034} & \colorbox{orange!25}{0.768} & \colorbox{red!25}{0.804} & \colorbox{red!25}{0.785} \\

\cmidrule[\heavyrulewidth]{1-12}
\end{tabular}
}
\end{table*}

%% file: tab/nvs.tex
\begin{table}[!t]
\centering
\caption{\textbf{Quantitative Comparison of Novel View Synthesis (NVS).} We evaluate NVS on both view extrapolation and interpolation set-ups, respectively.}
\setlength{\tabcolsep}{2pt} 
\renewcommand{\arraystretch}{1.2}
\begin{tabular}{c l ccc ccc}
\cmidrule[\heavyrulewidth]{1-8}
\multirow{2}{*}{} & \multirow{2}{*}{\textbf{Method}} & \multicolumn{3}{c}{\textbf{ScanNet++ (Extra.)}} & \multicolumn{3}{c}{\textbf{DeepBlending (Inter.)}} \\
\cmidrule(lr){3-5} \cmidrule(lr){6-8}
 & & PSNR $\uparrow$ & SSIM $\uparrow$ & LPIPS $\downarrow$ & PSNR $\uparrow$ & SSIM $\uparrow$ & LPIPS $\downarrow$ \\
\midrule

\multirow{5}{*}{\rotatebox{90}{Expl.}} 
 & 2DGS \cite{2DGS}    & 23.24   & 0.849   & 0.258   & \colorbox{yellow!25}{29.47}   & 0.903   & 0.256   \\
 & GOF \cite{GOF}      & \colorbox{orange!25}{23.54} & \colorbox{red!25}{0.865} & \colorbox{red!25}{0.237} & 29.36   & \colorbox{yellow!25}{0.905} & \colorbox{orange!25}{0.245} \\ 
 & RadeGS \cite{Rade-GS} & 23.27   & \colorbox{yellow!25}{0.858} & 0.244   & 29.16   & \colorbox{orange!25}{0.906} & \colorbox{yellow!25}{0.246}   \\ 
 & PGSR \cite{PGSR}    & 23.42   & 0.851   & 0.246   & 28.43   & 0.887   & 0.248   \\
 & GeoSVR \cite{GeoSVR} & 23.13   & 0.841   & 0.243   & 29.31   & 0.903   & 0.260   \\
\midrule

\multirow{3}{*}{\rotatebox{90}{Hybr.}} 
 & GSDF \cite{GSDF}    & \colorbox{yellow!25}{23.44}   & 0.857   & \colorbox{orange!25}{0.238} & \colorbox{red!25}{30.06} & \colorbox{red!25}{0.911} & \colorbox{red!25}{0.235} \\ 
 & GS-Pull \cite{GS-Pull} & 21.93   & 0.799   & 0.357   & 26.14   & 0.838   & 0.296   \\
 & \textbf{Ours}      & \colorbox{red!25}{23.76} & \colorbox{orange!25}{0.861} & \colorbox{yellow!25}{0.240}   & \colorbox{orange!25}{29.70} & 0.903   & 0.263   \\
\cmidrule[\heavyrulewidth]{1-8}
\end{tabular}
\label{tab:nvs}
\end{table}

%% file: tab/efficiency.tex


\begin{table}[t]
\begin{minipage}[t]{0.5\linewidth}
\centering
\caption{\textbf{Efficiency Analysis.} Time profiling is evaluated by averaging the total time costs of two selected scenes from ScanNet++ and DeepBlending, respectively.\label{tab:efficiency}}
\renewcommand{\arraystretch}{1.1}
\resizebox{\linewidth}{!}{%
\begin{tabular}{l ccc cc}
\cmidrule[\heavyrulewidth]{1-6}
\multirow{2}{*}{} & \multicolumn{3}{c}{\textbf{Explicit}} & \multicolumn{2}{c}{\textbf{Hybrid}} \\
\cmidrule(lr){2-4} \cmidrule(lr){5-6}
 & 2DGS & PGSR  & GeoSVR & GSDF  & Ours \\
\midrule
ScanNet++ \cite{scannet++}     & 9min  & 32min & 41min & $\approx3.5$h & 20min \\
DeepBlending \cite{DeepBlending}  & 21min & 55min & 31min & $\approx4$h & 26min \\
\cmidrule[\heavyrulewidth]{1-6}
\end{tabular}
}
\end{minipage}
\hfill 
\begin{minipage}[t]{0.45\linewidth}
\caption{\textbf{Ablation Study on ScanNet++.} Each design choice leads to a considerable performance gain across five comprehensive metrics.\label{tab:ablation}}
\renewcommand{\arraystretch}{1.1}
\resizebox{\linewidth}{!}
{
\begin{tabular}{l | c c c c c}
\cmidrule[\heavyrulewidth]{1-6}
Method & Prec $\uparrow$ & Recall $\uparrow$ & F-score $\uparrow$ & $D_{gs}$(cm) $\downarrow$ & $\delta_{5}$(\%) $\downarrow$ \\
\midrule
Anchor Sca.      & 0.519 & 0.946 & 0.659 & 7.0 & 14.53 \\
+ Voxel Sca. & 0.639 & 0.889 & 0.735 & /      & /     \\
+ Exp.       & 0.774 & 0.880 & 0.821 & 3.2 & 6.04  \\
+ Imp.        & \textbf{0.805} & \textbf{0.890} & \textbf{0.842} & \textbf{2.6} & \textbf{4.81} \\
\cmidrule[\heavyrulewidth]{1-6}
\end{tabular}
}
\end{minipage}
\end{table}

%% file: tab/ablation.tex

%% file: sec/5_conclusion.tex
\section{Conclusion and Limitation}
\label{sec:conclusion}
In this work, we present Gaussian-Voxel Duet, a dual-scaffolding hybrid representation designed for efficient and accurate surface reconstruction. With a 2D Gaussian anchor scaffold and a concurrently built sparse voxel grid along estimated surfaces, we further introduce explicit and implicit tethering strategies to facilitate their mutual regularization and co-optimization. As a result, structured Gaussian primitives are streamlined and lie close to the underlying surface defined by the SDF, greatly improving the compactness and efficacy in 3D reconstruction. In addition, we show superior novel view synthesis in extrapolated viewpoints owing to reliable geometry and much faster convergence compared to existing Gaussian-SDF hybrid representations.

\noindent \textbf{Limitation}. While our method demonstrates strong performance in monocular indoor reconstruction, we also provide preliminary and promising results on outdoor unbounded scenes in Appendix C.5. Nevertheless, we observe that the SDF inherently favors bounded, watertight surfaces, which are less representative of unbounded outdoor environments. Extending our dual-scaffold design with Unsigned Distance Field (UDF) offers a promising direction. In addition, we adopt a single-resolution voxel scaffold. Integrating a multi-resolution grid, as explored in Instant-NGP \cite{instant-ngp} and MonoSDF \cite{monosdf}, could further enhance the reconstruction of fine-grained geometric details.

%% file: sec/X_suppl.tex
\clearpage
\appendix
\setcounter{page}{1}

\begin{center}
    \vspace*{1em} 
    \Large \textbf{Supplementary Material for Gaussian-Voxel Duet}
    \vspace*{1em} 
\end{center}

\section{Overview}
\label{suppl:overview}
This supplementary material is organized as follows:
(1) Sec.~\ref{suppl:details} provides additional implementation details of our method, including training hyperparameters and mesh extraction.
(2) Sec.~\ref{suppl:experiments} presents details on the datasets, evaluation metrics, qualitative ablation studies, along with additional quantitative and qualitative experimental results.

\section{Implementation Details}
\label{suppl:details}

\subsection{Hyperparameters}

Most of our hyperparameters, including loss weights and optimization settings, are inherited from prior works~\cite{2DGS, scaffold-gs, ash}. Here, we summarize the key settings.

For the anchor-Gaussian objective in Eq.~\ref{eq:loss_gs}, we follow~\cite{2DGS, scaffold-gs} and set $\lambda_d = 10$ and $\lambda_n = 0.05$. The weights of the monocular depth and normal loss are initialized as $\lambda_{md} = 0.1$ and $\lambda_{mn} = 0.5$, and both are annealed to $0.01$ after 10,000 iterations.

For the voxel scaffold objective in Eq.~\ref{eq:loss_voxel}, we adopt the loss formulation of~\cite{ash} and set $\lambda_{vd} = 0.5$, $\lambda_{vn} = 0.1$, and $\lambda_{eik} = 0.1$. For the explicit tethering strategy in Eq.~\ref{eq:exp-tether}, we follow Scaffold-GS~\cite{scaffold-gs} and use $\tau_g = 0.0002$ and $\tau_p = 0.005$ as the gradient and opacity thresholds, respectively. The SDF threshold $\tau_d$ is chosen according to the scene scale: $0.1$ for ScanNetv2 and ScanNet++ scenes, and $0.3$ for DeepBlending scenes. For the implicit tethering loss in Eq.~\ref{eq:implicit-loss}, we set the associated loss weight to $\lambda_t = 5$. We optimize our model for a total of 20,000 iterations.

\subsection{Mesh Extraction}
We extract watertight meshes by directly applying Marching Cubes (MC) to the local SDF learned within our sparse voxel scaffold, requiring no additional post-processing.

\noindent\textbf{Explicit Gaussian methods.}
Prior explicit approaches~\cite{2DGS, GOF, Rade-GS, PGSR, GeoSVR} typically extract meshes via TSDF fusion of rendered depth maps. This multi-stage pipeline is computationally expensive and prone to noise and floaters due to multi-view inconsistencies.

\noindent\textbf{Implicit and hybrid SDF methods.}
Implicit~\cite{monosdf} and hybrid SDF-based methods~\cite{GSDF} learn a global SDF over the entire volume.
Standard evaluation protocols for these methods involve a multi-step process: first applying Marching Cubes to the learned SDF to generate a global mesh, then rendering depth maps from this mesh at the input camera poses, and finally re-fusing these depth maps via TSDF fusion to obtain the final mesh~\cite{manhattan-sdf, monosdf}.
While this procedure effectively restricts evaluation to observed regions, the unconstrained SDF in unobserved areas tends to generate spurious geometry and floaters. Furthermore, the ``MC $\rightarrow$ depth rendering $\rightarrow$ TSDF fusion'' pipeline significantly increases computational overhead.

\noindent\textbf{Ours.}
In contrast, our method allocates sparse voxels around surfaces and learns a local SDF, leaving free space unmeshed. This allows direct, high-quality mesh extraction via Marching Cubes, bypassing the time-consuming TSDF fusion steps required by baselines.

We also evaluate a variant using TSDF fusion on depth maps rendered from our tethered Gaussians, denoted as Ours (Fusion). As shown in Tab.~\ref{tab:mesh_extraction}, our direct extraction yields higher accuracy. Notably, even Ours (Fusion) outperforms the baselines, confirming that our tethering strategy aligns Gaussians closely with the underlying surfaces.

\begin{table}[!t]
\centering
\caption{\textbf{Surface Reconstruction on ScanNet++.}
\colorbox{red!25}{First}, \colorbox{orange!25}{second}, and \colorbox{yellow!25}{third} best results are highlighted.
MonoSDF and GSDF* use the standard ``MC $\rightarrow$ render $\rightarrow$ TSDF fusion'' protocol to obtain the mesh. Explicit methods (PGSR*, GeoSVR*) render depth maps and conduct TSDF fusion. In contrast, Ours directly extracts meshes from the local voxel SDF. Ours (Fusion) applies TSDF fusion to tethered Gaussians and already surpasses the baselines.}
\renewcommand{\arraystretch}{1.1}
\begin{tabular}{lccccc}
\toprule
Method        & Acc $\downarrow$  & Comp $\downarrow$  & Prec $\uparrow$ & Recall $\uparrow$ & F-score $\uparrow$ \\
\midrule
MonoSDF \cite{monosdf}   & 0.077 & 0.082 & 0.566 & 0.494 & 0.524 \\
GSDF* \cite{GSDF}         & \colorbox{orange!25}{0.071} & 0.113 & 0.612 & 0.516 & 0.559 \\
PGSR* \cite{PGSR}        & 0.081 & 0.047 & 0.718 & 0.732 & 0.723 \\
GeoSVR* \cite{GeoSVR}    & \colorbox{red!25}{0.046} & \colorbox{yellow!25}{0.044} & \colorbox{orange!25}{0.788} & \colorbox{yellow!25}{0.778} & \colorbox{yellow!25}{0.782} \\
Ours (Fusion)            & 0.083 & \colorbox{red!25}{0.022} & \colorbox{yellow!25}{0.704} & \colorbox{red!25}{0.936} & \colorbox{orange!25}{0.798} \\
Ours                     & \colorbox{yellow!25}{0.076} & \colorbox{orange!25}{0.027} & \colorbox{red!25}{0.805} & \colorbox{orange!25}{0.890} & \colorbox{red!25}{0.842} \\
\bottomrule
\end{tabular}
\label{tab:mesh_extraction}
\end{table}

\begin{figure*}[!t]
    \centering
    \includegraphics[width=1\linewidth]{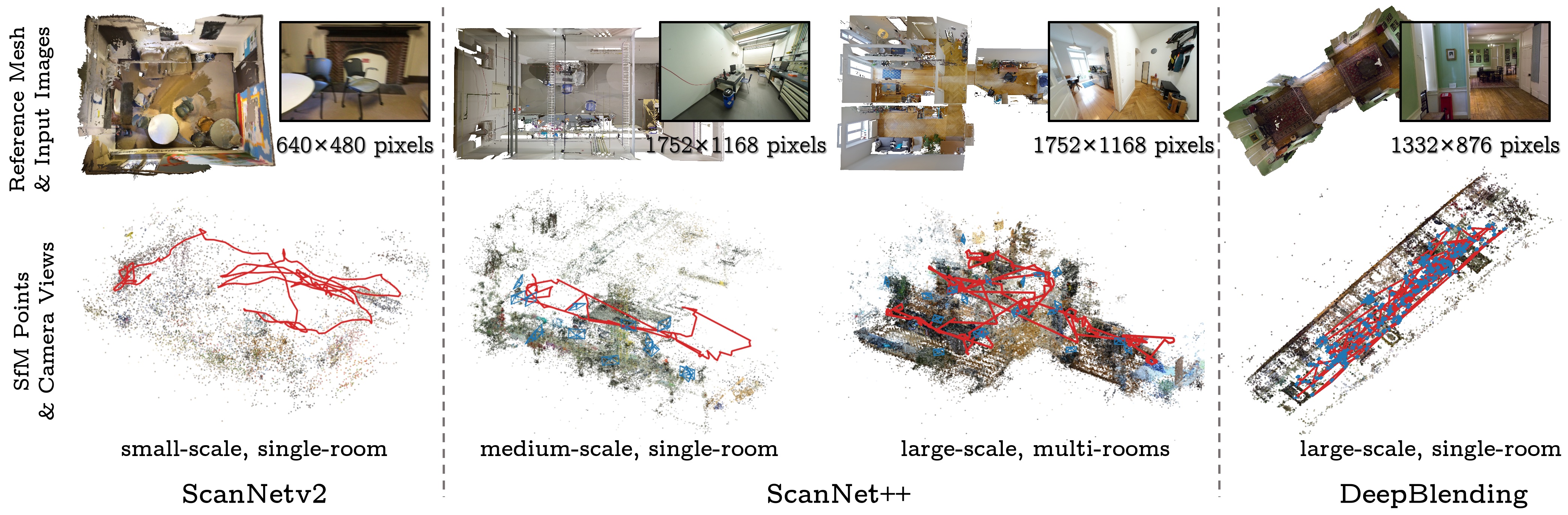}
    \caption{\textbf{Dataset Overview.} \textbf{ScanNetv2} contains small-scale single-room scenes with low-resolution, motion-blurred images and is used only for surface reconstruction; \textbf{ScanNet++} covers a range of scene scales and layout complexities with high-resolution DSLR images and is used for both surface reconstruction and challenging view-extrapolation NVS, where red and blue trajectories denote training and testing views, respectively; \textbf{DeepBlending} consists of large-scale scenes with high-resolution images and is used to evaluate NVS under a view-interpolation setup.}
    \label{fig:dataset}
\end{figure*}

\section{Experiments}
\label{suppl:experiments}

\begin{table}[!t]
\centering
\caption{
\textbf{Scene IDs Used in the Experiments.} In our experiments, we use 4 small-scale scenes from ScanNetv2~\cite{scannet}, 7 medium-scale scenes and 5 large-scale scenes from ScanNet++~\cite{scannet++}, and 2 large-scale scenes from DeepBlending~\cite{DeepBlending}. 
To evaluate our method under varying layout complexities, we include three multi-room ScanNet++ scenes (0a184cf634, 6f1848d1e3, 281ba69af1), which are marked with $^\dagger$ in the table; other scenes are single-room.}
\renewcommand{\arraystretch}{1.1}
\resizebox{0.9\linewidth}{!}{%
\begin{tabular}{c c l}
\toprule
\textbf{Dataset}      & \textbf{Scale}  & \textbf{Scene IDs} \\ 
\midrule
ScanNetv2 \cite{scannet} & small & scene0050\_00, scene0084\_00, scene0580\_00, scene0616\_00 \\
\midrule
\multirow{4}{*}{ScanNet++ \cite{scannet++}} 
             & \multirow{2}{*}{medium} 
                                 & 8b5caf3398, 0a184cf634$^\dagger$, 13c3e046d7, 1d003b07bd \\
             &                    & 260db9cf5a, 8be0cd3817, 6464461276 \\
 \cmidrule{2-3}
             & \multirow{2}{*}{large} 
                                 & 578511c8a9, 036bce3393, 6f1848d1e3$^\dagger$, 281ba69af1$^\dagger$ \\
             &                    & 9460c8889d \\
\midrule
DeepBlending \cite{DeepBlending} & large & drjohnson, playroom \\
\bottomrule
\end{tabular}}
\label{tab:scene_ids}
\end{table}

\subsection{Dataset Details}
We summarize the scenes used in our experiments in Tab.~\ref{tab:scene_ids}, which cover different spatial scales and layout complexities. Below, we provide additional details on image quality, resolution, and the evaluation setup for each dataset.

\noindent\textbf{ScanNet++.} 
We use the provided DSLR images at a resolution of $1752 \times 1168$ pixels, which offer high-quality, low-noise observations. ScanNet++ scenes are used to evaluate both surface reconstruction and novel view synthesis. 
Following the official split, ScanNet++ adopts a view extrapolation setup, where testing views deviate from the training trajectories, as illustrated in Fig.~\ref{fig:dataset}. 
This poses a much more challenging novel view synthesis setting than view interpolation, as models must generalize to viewpoints that lie outside the observed camera paths.

\noindent\textbf{DeepBlending.} 
We follow 3DGS~\cite{3DGS} and select the drjohnson and playroom scenes. The image resolutions are $1332 \times 876$ pixels for the drjohnson scene and $1264 \times 832$ pixels for the playroom scene. Due to the lack of reliable high-quality ground truth meshes in DeepBlending, we use this dataset exclusively for novel view synthesis evaluation. In contrast to ScanNet++, DeepBlending adopts a view interpolation setup, where testing views lie along the captured camera trajectories, leading to a more benign distribution of novel views.

\noindent\textbf{ScanNetv2.} 
We use all available images for each selected scene for training and focus on evaluating surface reconstruction performance on ScanNetv2. Following~\cite{manhattan-sdf, monosdf}, the input RGB images of ScanNetv2 scenes are resized to $640 \times 480$ pixels and often exhibit noticeable motion blur and sensor noise, resulting in relatively low image quality. We generate SfM point clouds from the given camera poses and images, but the degraded image quality leads to very sparse and noisy SfM points (see the first column of Fig.~\ref{fig:dataset}). In our ScanNetv2 experiments, we find that explicit methods tend to overfit to noisy observations, resulting in severe floaters and chaotic geometry. Hybrid baselines that use these optimized, yet noisy, Gaussian point clouds as priors for SDF learning inevitably propagate errors into the subsequent optimization process. In contrast, our method effectively resists such initial noise by dynamically pruning distant Gaussians and firmly tethering the remaining primitives to the learned SDF surface, preventing geometric drift and recovering clean structures.


\subsection{Evaluation Metrics}
For surface reconstruction, we follow the evaluation protocol of~\cite{atas} and compute accuracy (Acc), completeness (Comp), precision (Prec), recall (Recall), and F-score, with the definitions provided in Tab.~\ref{tab:metrics}.

\begin{table}[!htbp]
\centering
\caption{
\textbf{Surface Reconstruction Metrics.} $P$ and $P^*$ denote the point sets of the predicted and ground-truth meshes, respectively. The distance threshold is set to $T = 5$cm.}
\renewcommand{\arraystretch}{1.1}
\resizebox{0.5\linewidth}{!}{%
\begin{tabular}{l c}
\toprule
Metric   & Definition \\
\midrule
Acc      &
$\mathrm{mean}_{p \in P}\Bigl(\min_{p^* \in P^*} \lVert p - p^* \rVert_2\Bigr)$ \\

Comp     &
$\mathrm{mean}_{p^* \in P^*}\Bigl(\min_{p \in P} \lVert p - p^* \rVert_2\Bigr)$ \\

Prec     &
$\mathrm{mean}_{p \in P}\Bigl(\min_{p^* \in P^*} \lVert p - p^* \rVert_2 < T\Bigr)$ \\

Recall   &
$\mathrm{mean}_{p^* \in P^*}\Bigl(\min_{p \in P} \lVert p - p^* \rVert_2 < T\Bigr)$ \\

F-score  &
\small $\dfrac{2 \times \text{Prec} \times \text{Recall}}{\text{Prec} + \text{Recall}}$ \\
\bottomrule
\end{tabular}
}
\label{tab:metrics}
\end{table}

For novel view synthesis, we report standard PSNR, SSIM, and LPIPS between the rendered and ground-truth images, averaged over all test views.

To analyze the Gaussian point distribution in Sec.~\ref{sec:abla}, we additionally compute the mean distance $D_{gs}$ and the out-of-surface ratio $\delta_t$ of Gaussian centers. Let $P_{gs}$ be the point set of generated Gaussian centers and $P^*$ be the point set of the ground-truth mesh. $D_{gs}$ and $\delta_t$ are computed as:
\begin{equation}
    D_{gs} = \mathrm{mean}_{p_{gs} \in P_{gs}}
    \Bigl( \min_{p^* \in P^*} \lVert p_{gs} - p^* \rVert_2 \Bigr),
\end{equation}
\begin{equation}
    \delta_t = 
    \frac{\bigl|\{\,p_{gs} \in P_{gs} \mid \min_{p^* \in P^*} \lVert p_{gs} - p^* \rVert_2 > t \,\}\bigr|}{|P_{gs}|},
\end{equation}
where $t$ is a predefined distance threshold and $|\cdot|$ denotes the number of points in a set.

\begin{figure*}[!t]
    \centering
    \includegraphics[width=1\linewidth]{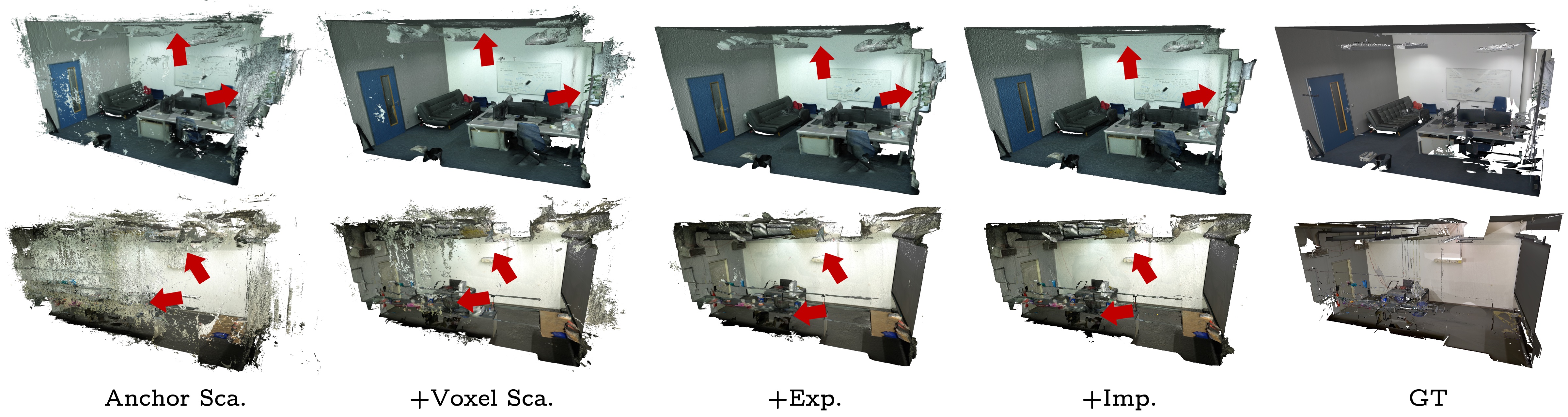}
    \caption{\textbf{Qualitative Results of the Ablation Study.} 
    The 2DGS anchor scaffold (Anchor Sca.) exhibits significant floating artifacts (indicated by red arrows). 
    Adding the voxel scaffold (Voxel Sca.) to learn a local SDF effectively eliminates this noise, producing clean geometry. 
    Incorporating explicit anchor tethering (Exp.) and implicit surface tethering (Imp.) further enhances surface details and completeness.}
    \label{fig:abla}
\end{figure*}

\subsection{Qualitative Results of Ablation Study}
We visually verify the effectiveness of our key design choices in Fig.~\ref{fig:abla}. The associated quantitative results can be seen in Tab.~\ref{tab:ablation}. As highlighted by the red arrows, the 2DGS anchor scaffold (Anchor Sca.) suffers from severe floating artifacts in free space. By introducing the voxel scaffold (Voxel Sca.), our method leverages local SDF learning to effectively suppress these floaters, yielding a significantly cleaner geometry. Finally, the successive application of explicit anchor tethering (Exp.) and implicit surface tethering (Imp.) further refines surface fidelity and recovers structural details, resulting in high-quality meshes that align closely with the ground truth.

\subsection{Comparison with Feed-Forward Methods}
\begin{figure*}[!t]
    \centering
    \includegraphics[width=0.8\linewidth]{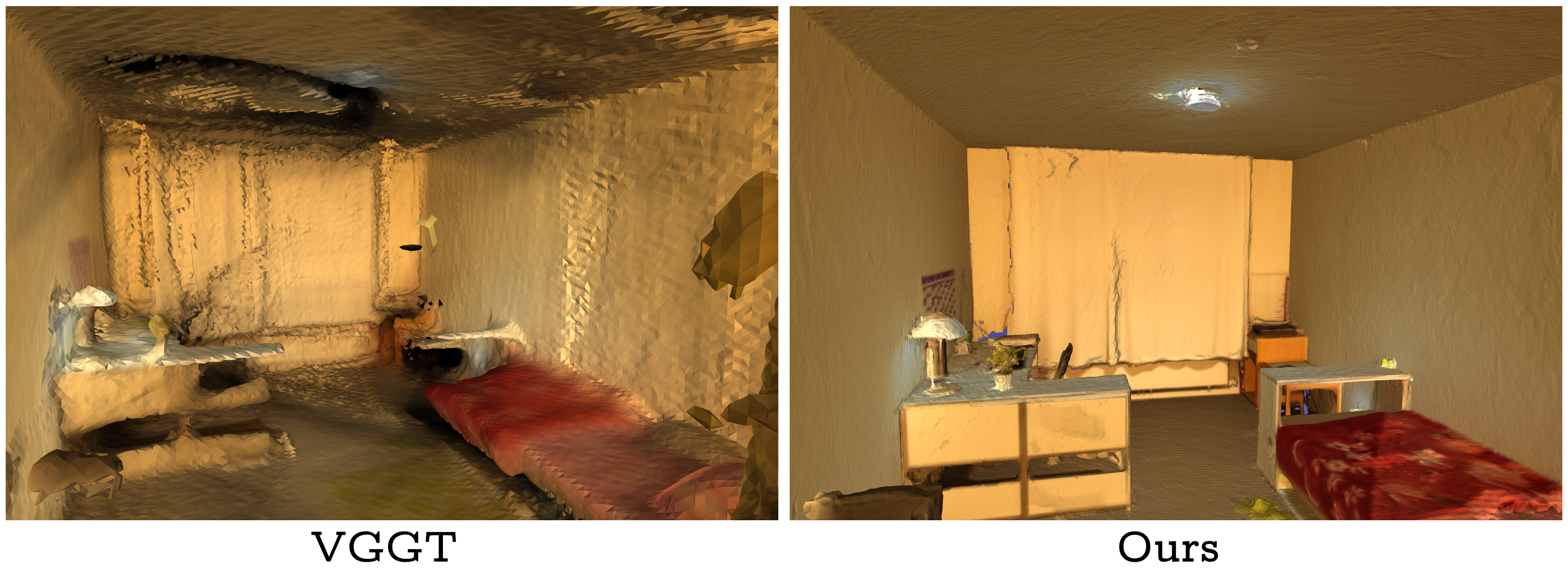}
    \caption{\textbf{Qualitative Comparison with VGGT.} 
    While VGGT enables fast single-pass inference, it yields coarse geometry. In contrast, our per-scene optimization remains essential for achieving high-fidelity reconstruction.}
    \label{fig:compare_vggt}
\end{figure*}

We also evaluate our method against VGGT, a recent generalizable feed-forward approach. By feeding all training images into VGGT, we extract depth maps and camera poses, which are subsequently processed via TSDF fusion to generate a global mesh. As shown in Fig.~\ref{fig:compare_vggt}, while VGGT enables fast single-pass inference, it yields coarse geometry with limited precision despite being pre-trained on large-scale data. In contrast, our per-scene optimization remains essential for capturing intricate details and achieving high-fidelity reconstruction. Consequently, bridging these two paradigms—by employing the output of feed-forward networks as a geometric initialization to accelerate our per-scene optimization—is a compelling direction for future work.

\subsection{Reconstruction Results on Objects and Outdoor Scenes}
\begin{figure*}[!t]
    \centering
    \includegraphics[width=1\linewidth]{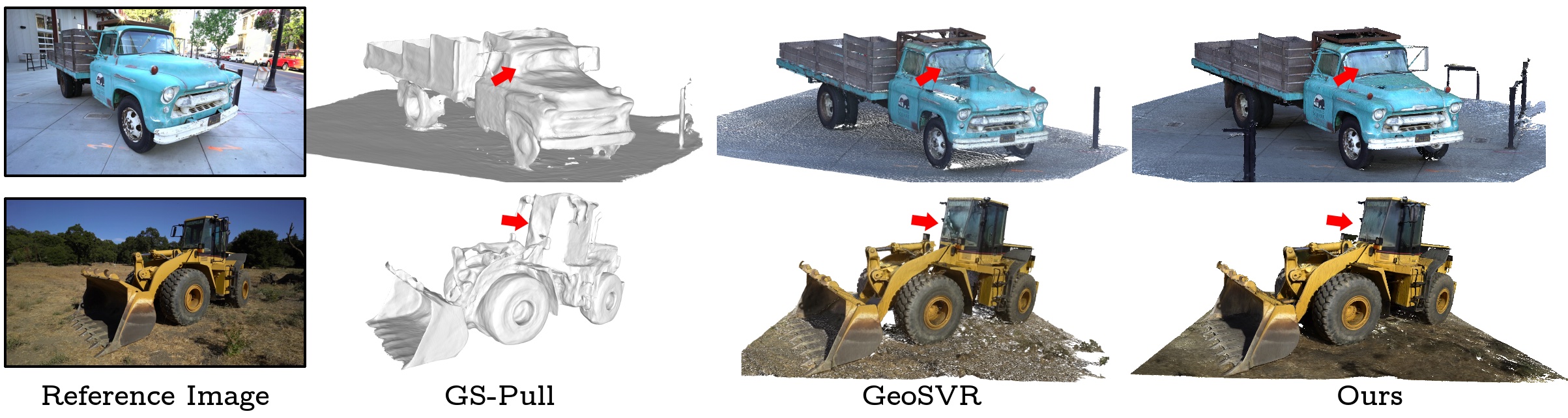}
    \caption{\textbf{Qualitative Comparison on TNT.} Compared to baselines like GS-Pull and GeoSVR, our method demonstrates enhanced robustness when handling reflective surfaces.}
    \label{fig:tnt}
\end{figure*}

\input{tab/dtu_tnt}

While our method focuses on monocular indoor reconstruction, we also evaluate it on the standard DTU and TNT benchmarks. As illustrated in Tab. \ref{tab:dtu_tnt}, our approach outperforms the hybrid baselines GSDF and GS-Pull. Fig. \ref{fig:tnt} illustrates that our method exhibits greater robustness on challenging reflective surfaces, such as window glass. However, we observe that explicit methods like GeoSVR still lead on these datasets, likely because SDF is advantageous for bounded indoor layouts with watertight objects, rather than open surfaces in unbounded scenes. As discussed in the limitations, integrating multi-resolution grid and Unsigned Distance Field (UDF) into our voxel-scaffold representation offers a promising direction to further enhance reconstruction quality.

\subsection{Additional Results}
\begin{figure*}[!t]
    \centering
    \includegraphics[width=1\linewidth]{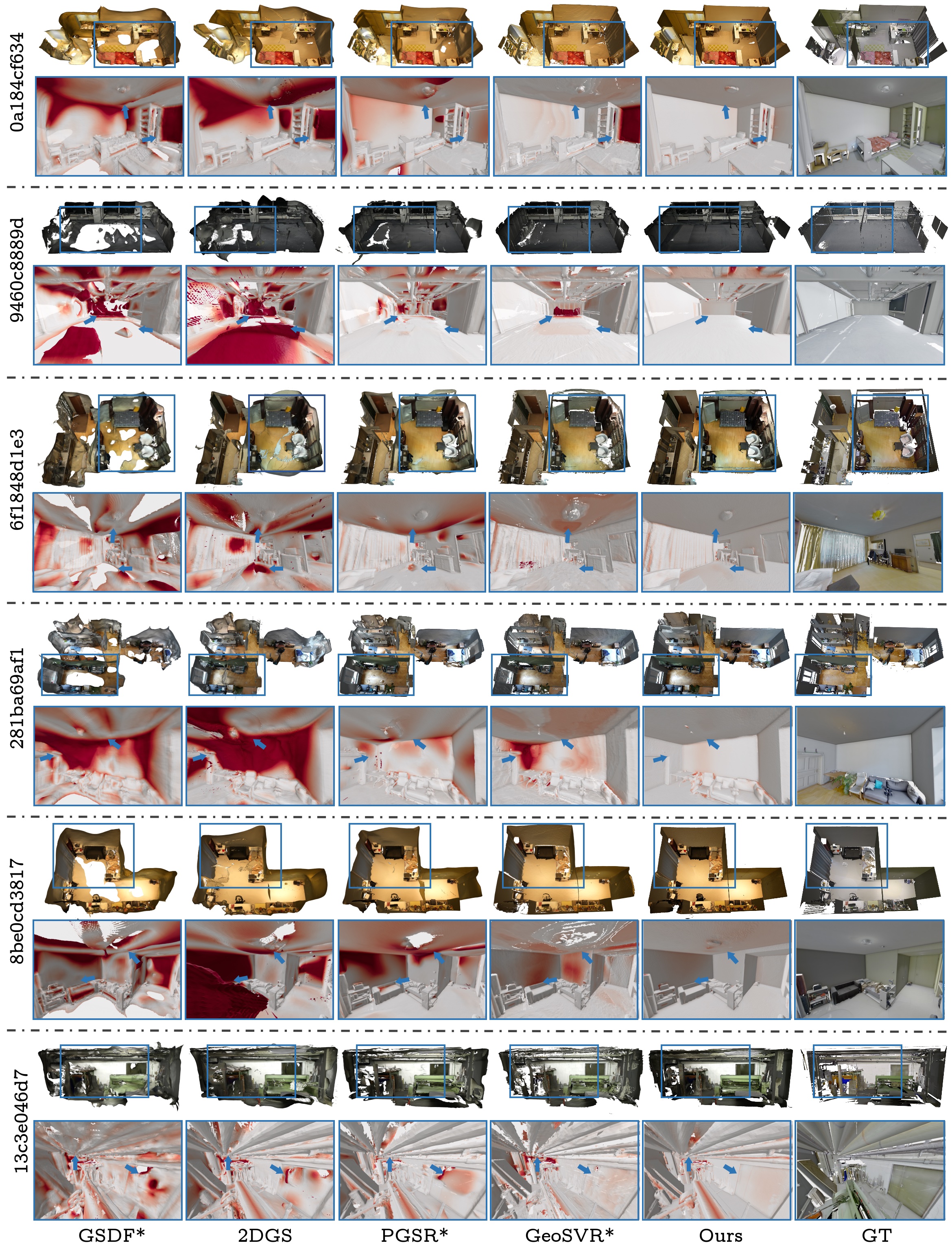}
    \caption{\textbf{Qualitative Results of Surface Reconstruction on ScanNet++.} 
    We show the global mesh (top) and error map of the region indicated by the blue box (bottom) of six scenes. Reconstruction errors are visualized from white (low) to red (high). While 2DGS and PGSR* struggle in textureless areas, GSDF* fails to learn an accurate global SDF in large-scale scenes, and GeoSVR* produces visible holes. In contrast, our method achieves superior geometric accuracy.
    }
    \label{fig:surface_recon_scannetpp_suppl}
\end{figure*}

\begin{figure*}[!t]
    \centering
    \includegraphics[width=1\linewidth]{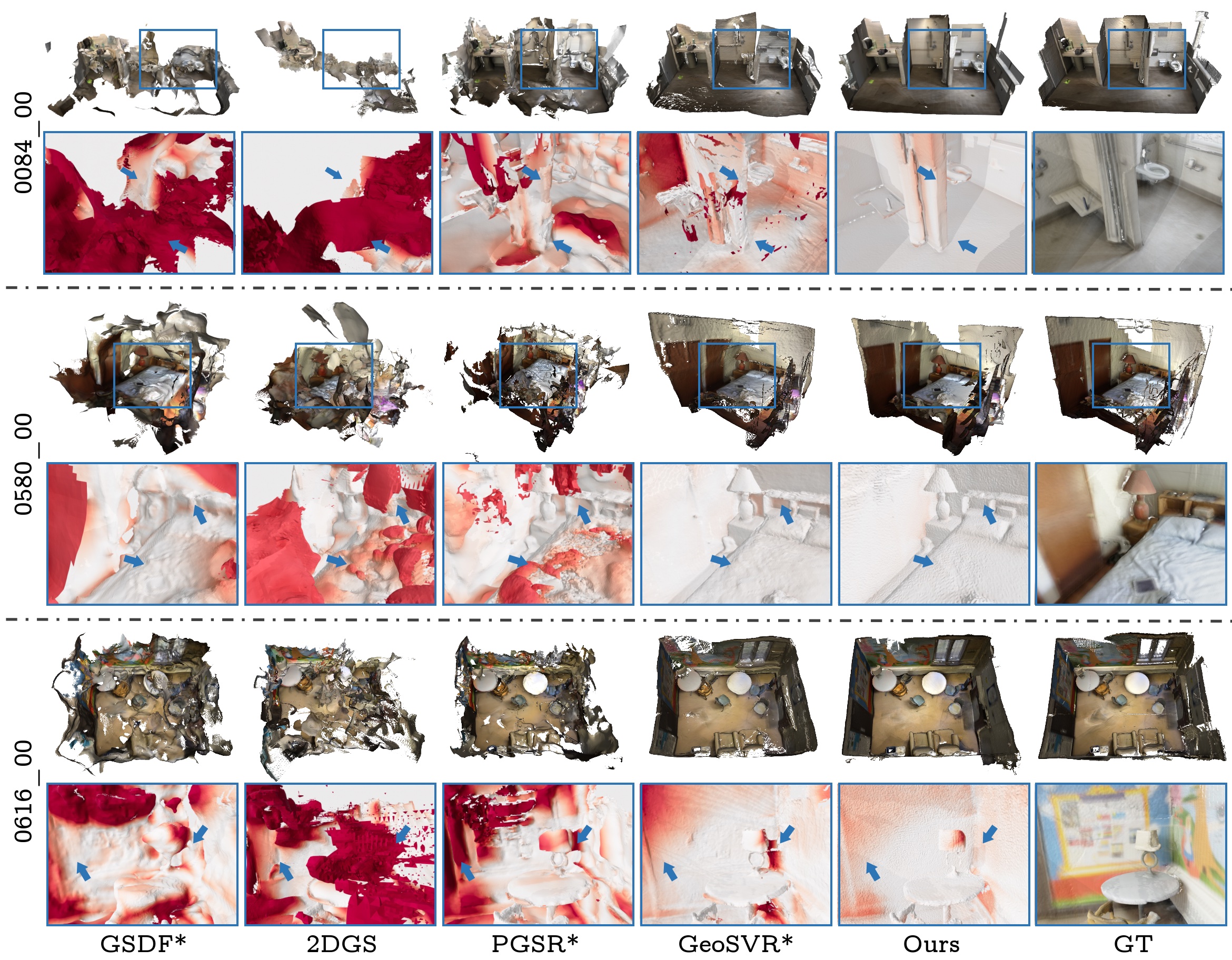}
    \caption{\textbf{Qualitative Results of Surface Reconstruction on ScanNetv2.} 
    We show the global mesh (top) and error map of the region indicated by the blue box (bottom) of three scenes. Existing baselines like 2DGS, PGSR*, and GSDF* suffer from degraded image quality and noisy SfM initializations, leading to poor surface reconstruction. Although GeoSVR* attempts to recover high-frequency details, the absence of planar constraints results in noisy and uneven surface. In contrast, our method achieves a favorable balance between preserving fine details and ensuring surface smoothness, yielding the most accurate and structurally coherent reconstructions.
    }
    \label{fig:surface_recon_scannet_suppl}
\end{figure*}

We further present extended qualitative results of surface reconstruction on the ScanNet++ and ScanNetv2 datasets in Fig.~\ref{fig:surface_recon_scannetpp_suppl} and Fig.~\ref{fig:surface_recon_scannet_suppl}, respectively. To highlight the advantages of our approach, we visualize the error maps of the reconstructed meshes, which clearly demonstrate our superior accuracy. As shown in Fig.~\ref{fig:surface_recon_scannetpp_suppl}, existing Gaussian-based methods such as 2DGS and PGSR* struggle in textureless regions and often fail to recover valid geometry. While GeoSVR* utilizes voxel constraints, it tends to produce incomplete meshes with significant holes, resulting in poor coverage of the underlying scene. In contrast, our method generates high-quality 3D meshes with complete and faithful geometry.

On ScanNetv2, explicit methods typically undergo significant performance degradation due to challenges such as low image resolution, motion blur, and noisy SfM initializations. However, our method exhibits strong robustness, effectively suppressing noise and recovering clean surfaces even under these adverse conditions.

\begin{figure*}[!t]
    \centering
    \includegraphics[width=1\linewidth]{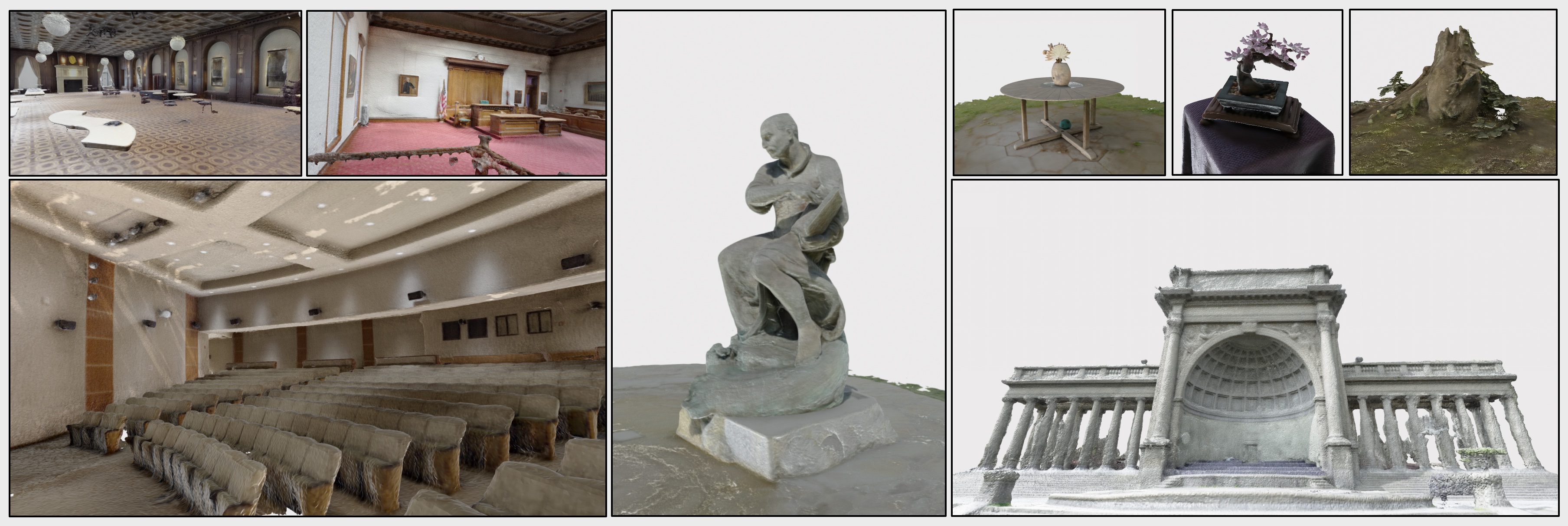}
    \caption{\textbf{Extended Results on Large-Scale Scenes.} Our method successfully scales to diverse, expansive scenes.} 
    \label{fig:surface_recon_demo}
\end{figure*}

In addition to qualitative comparisons, we report detailed scene-wise quantitative results of our experiments in Tab. \ref{tab:stat_1}, \ref{tab:stat_2}, \ref{tab:stat_3}, and \ref{tab:stat_4}. Furthermore, Fig.~\ref{fig:surface_recon_demo} illustrates the scalability of our method by showcasing reconstruction results for very large-scale scenes.

\input{tab/statistics}

%% file: tab/dtu_tnt.tex
\begin{table}[!t]
\centering
\caption{\textbf{Quantitative Results on DTU and TNT.} 
Quantitative metrics of the baselines are sourced from their original papers. As shown, our approach achieves superior performance compared to existing hybrid baselines.
}
\renewcommand{\arraystretch}{1.1}
\resizebox{0.6\linewidth}{!}
{%
\setlength{\tabcolsep}{5pt}
\begin{tabular}{c|c|ccc}
\toprule
    \multirow{2}{*}{\textbf{Method}} & \textbf{Expl}. & \multicolumn{3}{c}{\textbf{Hybr}.} \\
    \cmidrule(lr){2-2} \cmidrule(lr){3-5}
    & GeoSVR & GSDF & GS-Pull & Ours \\
    \midrule
    DTU (CD $\downarrow$) & 0.47 & 0.80 & 0.75 & \textbf{0.55} \\
    TNT (F-score $\uparrow$) & 0.56 & - & 0.43 & \textbf{0.50} \\
\bottomrule
\end{tabular}
}
\label{tab:dtu_tnt}
\end{table}

%% file: tab/statistics.tex
\begin{table*}[!htbp]
    \centering
    \caption{\textbf{Scene-wise Quantitative Results of Surface Reconstruction on ScanNet++.} We report Accuracy (Acc), Completeness (Comp), Precision (Prec), Recall (Rec), and F-score. For Acc and Comp, lower is better ($\downarrow$); for others, higher is better ($\uparrow$). Best results are highlighted in \textbf{bold}.}
    \label{tab:scannetpp_per_scene}
    \resizebox{\linewidth}{!}{%
        \begin{tabular}{l|ccccc|ccccc|ccccc}
            \toprule
            \multirow{2}{*}{Scene} & \multicolumn{5}{c|}{PGSR$^\ast$} & \multicolumn{5}{c|}{GeoSVR$^\ast$} & \multicolumn{5}{c}{Ours} \\
             & Acc $\downarrow$ & Comp $\downarrow$ & Prec $\uparrow$ & Rec $\uparrow$ & F-score $\uparrow$ 
              & Acc $\downarrow$ & Comp $\downarrow$ & Prec $\uparrow$ & Rec $\uparrow$ & F-score $\uparrow$ 
             & Acc $\downarrow$ & Comp $\downarrow$ & Prec $\uparrow$ & Rec $\uparrow$ & F-score $\uparrow$ \\
            \midrule
            0a184cf634 & 0.040 & 0.041 & 0.715 & 0.704 & 0.710  & 0.032 & 0.039 & 0.855 & 0.836 & 0.845 & \textbf{0.025} & \textbf{0.024} & \textbf{0.903} & \textbf{0.923} & \textbf{0.913} \\
            036bce3393 & 0.034 & 0.038 & 0.783 & 0.781 & 0.782  & 0.032 & 0.036 & 0.836 & 0.818 & 0.827 & \textbf{0.029} & \textbf{0.027} & \textbf{0.837} & \textbf{0.872} & \textbf{0.854} \\
            13c3e046d7 & \textbf{0.031} & 0.047 & \textbf{0.816} & 0.777 & 0.796  & 0.035 & 0.066 & 0.815 & 0.737 & 0.774 & 0.039 & \textbf{0.035} & 0.764 & \textbf{0.834} & \textbf{0.798} \\
            1d003b07bd & 0.037 & 0.042 & 0.778 & 0.784 & 0.781  & 0.035 & 0.038 & 0.801 & 0.799 & 0.800 & \textbf{0.032} & \textbf{0.024} & \textbf{0.839} & \textbf{0.921} & \textbf{0.878} \\
            260db9cf5a & 0.045 & 0.048 & 0.760 & 0.749 & 0.754  & 0.033 & 0.031 & 0.809 & 0.821 & 0.815 & \textbf{0.028} & \textbf{0.019} & \textbf{0.872} & \textbf{0.950} & \textbf{0.909} \\
            281ba69af1 & 0.056 & 0.073 & 0.618 & 0.559 & 0.587  & \textbf{0.042} & 0.061 & 0.754 & 0.667 & 0.708 & 0.045 & \textbf{0.040} & \textbf{0.767} & \textbf{0.776} & \textbf{0.772} \\
            8be0cd3817 & 0.048 & 0.040 & 0.731 & 0.724 & 0.727  & 0.026 & 0.026 & 0.860 & 0.863 & 0.862 & \textbf{0.018} & \textbf{0.017} & \textbf{0.954} & \textbf{0.969} & \textbf{0.961} \\
            8b5caf3398 & 0.161 & 0.031 & 0.628 & 0.823 & 0.712  & \textbf{0.052} & 0.035 & \textbf{0.715} & 0.799 & 0.754 & 0.190 & \textbf{0.021} & 0.643 & \textbf{0.961} & \textbf{0.770} \\
            6464461276 & 0.045 & 0.039 & 0.752 & 0.766 & 0.759  & 0.034 & 0.040 & 0.757 & 0.755 & 0.756 & \textbf{0.030} & \textbf{0.024} & \textbf{0.867} & \textbf{0.900} & \textbf{0.883} \\
            578511c8a9 & 0.282 & 0.042 & 0.676 & 0.774 & 0.722  & \textbf{0.057} & 0.055 & \textbf{0.678} & 0.700 & 0.689 & 0.303 & \textbf{0.025} & 0.645 & \textbf{0.903} & \textbf{0.752} \\
            6f1848d1e3 & 0.130 & 0.063 & 0.681 & 0.649 & 0.664  & \textbf{0.123} & 0.046 & 0.790 & 0.772 & 0.781 & \textbf{0.123} & \textbf{0.039} & \textbf{0.801} & \textbf{0.813} & \textbf{0.807} \\
            9460c8889d & 0.062 & 0.057 & 0.678 & 0.692 & 0.685  & \textbf{0.051} & 0.053 & \textbf{0.784} & 0.774 & 0.779 & 0.055 & \textbf{0.029} & 0.764 & \textbf{0.859} & \textbf{0.809} \\
            \midrule
            Average & 0.081 & 0.047 & 0.718 & 0.732 & 0.723  & \textbf{0.046} & 0.044 & 0.788 & 0.778 & 0.782 & 0.076 & \textbf{0.027} & \textbf{0.805} & \textbf{0.890} & \textbf{0.842} \\
            \bottomrule
        \end{tabular}%
    }
\label{tab:stat_1}
\end{table*}

\begin{table*}[!htbp]
    \centering
    \caption{\textbf{Scene-wise Quantitative Results of Surface Reconstruction on ScanNetv2.} We report Accuracy (Acc), Completeness (Comp), Precision (Prec), Recall (Rec), and F-score. For Acc and Comp, lower is better ($\downarrow$); for others, higher is better ($\uparrow$). Best results are highlighted in \textbf{bold}.}
    \label{tab:scannet_per_scene}
    \resizebox{\linewidth}{!}{%
        \begin{tabular}{l|ccccc|ccccc|ccccc}
            \toprule
            \multirow{2}{*}{Scene} & \multicolumn{5}{c|}{PGSR$^\ast$} & \multicolumn{5}{c|}{GeoSVR$^\ast$} & \multicolumn{5}{c}{Ours} \\
             & Acc $\downarrow$ & Comp $\downarrow$ & Prec $\uparrow$ & Rec $\uparrow$ & F-score $\uparrow$ 
              & Acc $\downarrow$ & Comp $\downarrow$ & Prec $\uparrow$ & Rec $\uparrow$ & F-score $\uparrow$ 
             & Acc $\downarrow$ & Comp $\downarrow$ & Prec $\uparrow$ & Rec $\uparrow$ & F-score $\uparrow$ \\
            \midrule
            scene0050\_00 & 0.122 & 0.118 & 0.480 & 0.471 & 0.475  & 0.046 
& 0.051 & 0.689 & 0.633 & 0.660 & \textbf{0.045} & \textbf{0.030} & \textbf{0.818} & \textbf{0.844} & \textbf{0.831} \\
            scene0084\_00 & 0.116 & 0.091 & 0.343 & 0.353 & 0.348  & \textbf{0.082}& 0.082 & 0.418 & 0.447 & 0.432 & 0.090 & \textbf{0.029} & \textbf{0.780} & \textbf{0.871} & \textbf{0.823} \\
            scene0580\_00 & 0.137 & 0.213 & 0.442 & 0.379 & 0.408  & 0.058 
& 0.062 & 0.601 & 0.575 & 0.587 & \textbf{0.043} & \textbf{0.037} & \textbf{0.733} & \textbf{0.750} & \textbf{0.741} \\
            scene0616\_00 & 0.131 & 0.144 & 0.319 & 0.283 & 0.300  & 0.057 
& 0.061 & 0.592 & 0.551 & 0.571 & \textbf{0.042} & \textbf{0.037} & \textbf{0.741} & \textbf{0.753} & \textbf{0.747} \\
            \midrule
            Average & 0.127 & 0.141 & 0.396 & 0.372 & 0.383  & 0.061 
& 0.064 & 0.575 & 0.551 & 0.562 & \textbf{0.055}& \textbf{0.034} & \textbf{0.768} & \textbf{0.804} & \textbf{0.785} \\
            \bottomrule
        \end{tabular}%
    }
\label{tab:stat_2}
\end{table*}

\begin{table*}[!htbp]
    \centering
    \caption{\textbf{Scene-wise Quantitative Results of Novel View Synthesis on ScanNet++.} We report PSNR, SSIM, and LPIPS. For PSNR and SSIM, higher is better ($\uparrow$); for LPIPS, lower is better ($\downarrow$). Best results are highlighted in \textbf{bold}.}
    \label{tab:nvs_scannetpp}
    \resizebox{0.8\linewidth}{!}{%
        \begin{tabular}{l|ccc|ccc|ccc}
            \toprule
            \multirow{2}{*}{Scene} & \multicolumn{3}{c|}{PGSR} & \multicolumn{3}{c|}{GeoSVR} & \multicolumn{3}{c}{Ours} \\
             & PSNR $\uparrow$ & SSIM $\uparrow$ & LPIPS $\downarrow$ 
              & PSNR $\uparrow$ & SSIM $\uparrow$ & LPIPS $\downarrow$ 
             & PSNR $\uparrow$ & SSIM $\uparrow$ & LPIPS $\downarrow$ \\
            \midrule
            036bce3393 & 25.36 & \textbf{0.851} & \textbf{0.195}  & 23.80 & 0.808 & 0.232 & \textbf{25.41} & 0.849 & 0.220 \\
            0a184cf634 & \textbf{26.80} & 0.905 & 0.201  & 26.67 & 0.901 & \textbf{0.194} & 26.67 & \textbf{0.907} & 0.212 \\
            13c3e046d7 & 21.90 & 0.823 & 0.294  & \textbf{22.02} & 0.812 & 0.285 & 21.88 & \textbf{0.835} & \textbf{0.284} \\
            1d003b07bd & 20.16 & 0.795 & 0.295  & 19.98 & 0.802 & 0.264 & \textbf{20.17} & \textbf{0.814} & \textbf{0.258} \\
            260db9cf5a & 26.17 & 0.885 & \textbf{0.179}  & 25.00 & 0.860 & 0.194 & \textbf{26.79} & \textbf{0.891} & 0.184 \\
            281ba69af1 & 19.00 & \textbf{0.835} & 0.285  & \textbf{19.51} & 0.831 & \textbf{0.277} & 18.13 & 0.818 & 0.285 \\
            578511c8a9 & 21.86 & 0.831 & \textbf{0.246}  & 20.96 & 0.808 & 0.261 & \textbf{22.63} & \textbf{0.846} & 0.251 \\
            6464461276 & 21.61 & 0.812 & 0.270  & 21.73 & 0.803 & 0.260 & \textbf{22.08} & \textbf{0.824} & \textbf{0.254} \\
            6f1848d1e3 & 22.88 & 0.863 & 0.271  & 22.38 & 0.850 & \textbf{0.254} & \textbf{23.44} & \textbf{0.874} & 0.272 \\
            8b5caf3398 & 27.06 & 0.904 & 0.171  & 25.96 & 0.889 & 0.189 & \textbf{27.57} & \textbf{0.913} & \textbf{0.169} \\
            8be0cd3817 & 27.35 & \textbf{0.910} & 0.218  & \textbf{27.77} & 0.901 & \textbf{0.213} & 27.63 & 0.908 & 0.226 \\
            9460c8889d & 20.84 & 0.802 & 0.323  & 21.74 & 0.823 & 0.292 & \textbf{22.72} & \textbf{0.851} & \textbf{0.267} \\
            \midrule
            Average & 23.42 & 0.851 & 0.246  & 23.13 & 0.841 & 0.243 & \textbf{23.76} & \textbf{0.861} & \textbf{0.240} \\
            \bottomrule
        \end{tabular}%
    }
\label{tab:stat_3}
\end{table*}

\begin{table*}[!htbp]
    \centering
    \caption{\textbf{Scene-wise Quantitative Results of Novel View Synthesis on DeepBlending.} We report PSNR, SSIM, and LPIPS. For PSNR and SSIM, higher is better ($\uparrow$); for LPIPS, lower is better ($\downarrow$). Best results are highlighted in \textbf{bold}.}
    \label{tab:nvs_db}
    \resizebox{0.8\linewidth}{!}{%
        \begin{tabular}{l|ccc|ccc|ccc}
            \toprule
            \multirow{2}{*}{Scene}  & \multicolumn{3}{c|}{PGSR}  & \multicolumn{3}{c|}{GeoSVR} & \multicolumn{3}{c}{Ours} \\
              & PSNR $\uparrow$ & SSIM $\uparrow$ & LPIPS $\downarrow$ & PSNR $\uparrow$ & SSIM $\uparrow$ & LPIPS $\downarrow$ & PSNR $\uparrow$ & SSIM $\uparrow$ & LPIPS $\downarrow$ \\
            \midrule
            drjohnson & 27.06 & 0.868 & \textbf{0.265} & \textbf{28.82} & \textbf{0.902} & 0.300 & 28.78 & 0.898 & 0.268 \\
            playroom & 29.81 & 0.906 & 0.231 & 29.80 & 0.903 & \textbf{0.220} & \textbf{30.63} & \textbf{0.908} & 0.258 \\
            \midrule
            Average & 28.43 & 0.887 & \textbf{0.248} & 29.31 & \textbf{0.903} & 0.260 & \textbf{29.70} & \textbf{0.903} & 0.263 \\
            \bottomrule
        \end{tabular}%
    }
    \label{tab:stat_4}
\end{table*}